\documentclass[10pt,twocolumn,letterpaper]{article}

\usepackage[pagenumbers]{wacv} 

\definecolor{wacvblue}{rgb}{0.21,0.49,0.74}
\usepackage[pagebackref,breaklinks,colorlinks,allcolors=wacvblue]{hyperref}

\def\wacvPaperID{1433} 
\def\confName{WACV}
\def\confYear{2027}

\title{Imprint: Online Memory Compression for Long-Horizon Egocentric QA}

\author{Kousik Das\\
IIT, Kharagpur\\
India\\
{\tt\small kousik.24@kgpian.iitkgp.ac.in}
\and
Debaditya Roy\\
IIT, Kharagpur\\
India\\
{\tt\small debaditya@cse.iitkgp.ac.in}
}

\usepackage[T1]{fontenc}
\usepackage{booktabs}
\usepackage{float}
\usepackage{amsmath} 
\usepackage{xcolor} 
\usepackage{algorithm}
\usepackage{algpseudocode}
\usepackage{amsfonts}
\usepackage{makecell}
\usepackage{xcolor}
\usepackage{comment} 
\usepackage{printlen}

\usepackage[normalem]{ulem} 
\usepackage{amsmath}
\usepackage{amssymb}
\usepackage{bm}

\usepackage{tcolorbox}
\usepackage{listings}
\usepackage{courier}
\begin{document}
\maketitle
\begin{abstract}
Long-horizon egocentric question answering involves answering about events that have occurred hours or days in the past. This requires memory representations that remain both retrieval-effective and scalable over days or weeks of recording. Existing long-horizon egocentric QA methods construct memory as hierarchical textual summaries of observations. While effective for reducing memory size, summarization optimizes for descriptive compression rather than retrieval: repeated interactions are absorbed into coarse textual descriptions instead of being preserved as explicit, recurring memory units, making long-horizon evidence aggregation difficult. We propose Imprint, an interaction-centric memory framework that formulates long-horizon egocentric memory as an online memory compression problem rather than  summarization. Incoming observations are first represented as structured Interaction Records and continuously organized into recurring interaction patterns. Using human memory consolidation signals of recurrence, recency, and distinctiveness, Imprint selectively retains and compresses interactions into a compact retrieval-oriented memory. We evaluate Imprint on EgoLifeQA, a seven-day egocentric benchmark containing questions that require reasoning over interactions occurring hours to days before the query. With the same LLM, Imprint improves QA accuracy from 31.0\% to 35.8\%, increases evidence-grounded answers by $6\times$ compared with EgoRAG, reduces memory footprint by $2.3\times$, and decreases retrieval latency by $11.8\times$. These results demonstrate that memory compression provides a  scalable and retrieval-effective foundation for long-horizon egocentric question answering.

\end{abstract}    
\section{Introduction}
\label{sec:Introduction}
\begin{figure}[!htbp]
  \centering
  \includegraphics[width=\columnwidth]{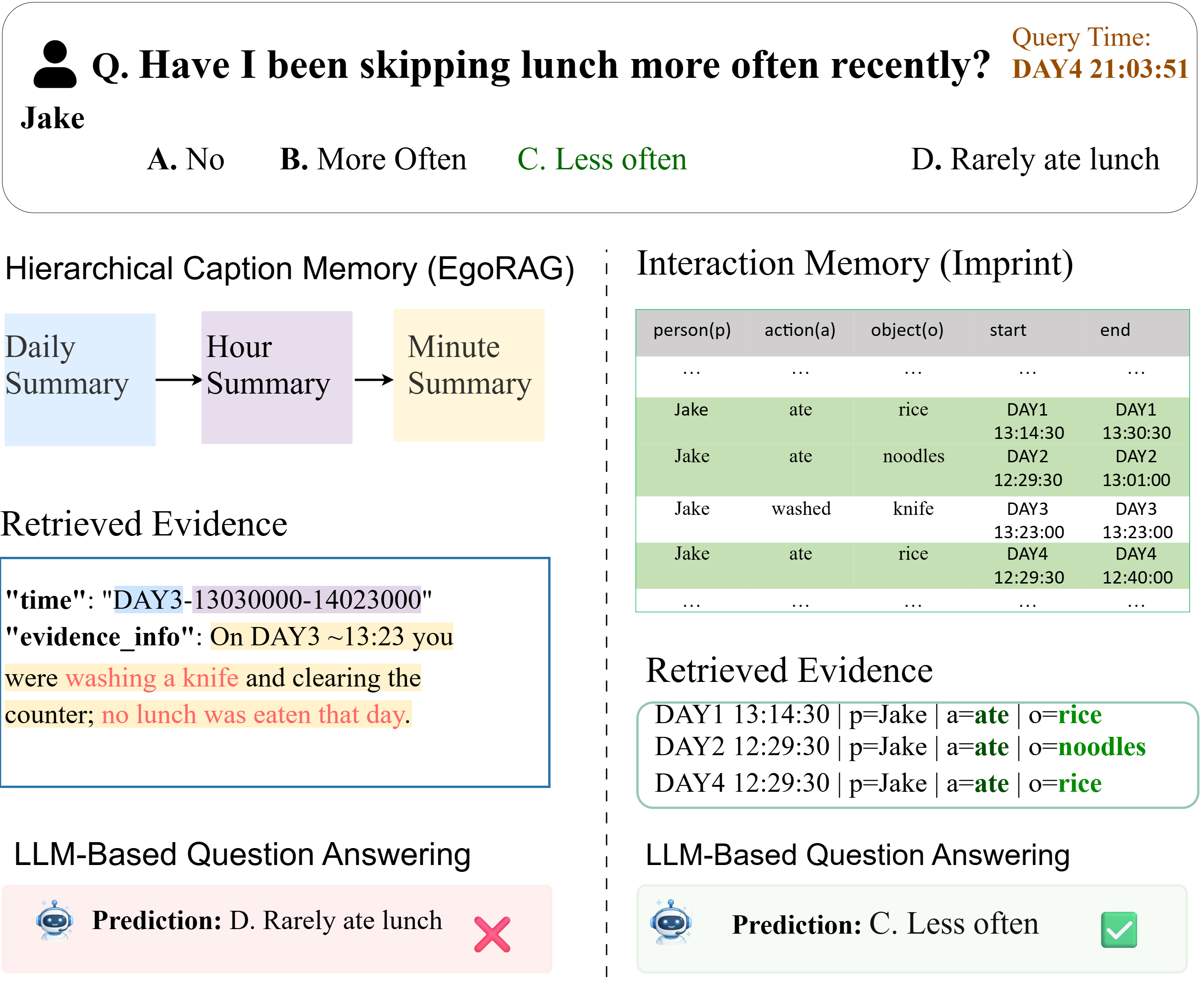}
  \caption{Interaction structure matters. Imprint aggregates meal interactions across days to find the correct trend (\textit{Less often}), whereas caption-based memory (e.g. EgoRAG~\cite{yang2025egolife}) retrieves only same-day evidence (\textit{no lunch on DAY3}) and predicts incorrectly (\textit{Rarely ate lunch}). Both methods use the same \textsc{Qwen2.5-7B-Instruct} model for question answering.}
  \label{fig:teaser_fig}
\end{figure}
Long-term egocentric memory assistants enable users to answer questions about interactions captured through wearable cameras, supporting applications such as autobiographical memory assistance~\cite{subramaniam2026detecting}, personal analytics, and cognitive support. Unlike conventional video understanding~\cite{zhang2023video, maaz2023video, li2024llava}, which answers over seconds to minutes of videos, egocentric memory assistants must answer queries about events that have occurred hours or days before. To enable efficient retrieval over such long videos, existing methods \cite{yang2025egolife,tian2026ego} first transform raw video into unstructured captions and then into hierarchical textual summaries. 

While effective for describing individual observations and reducing memory volume compared to videos, identify recurring person--action--object patterns and aggregating evidence across long time horizons is challenging in textual summaries. This limitation becomes particularly apparent for behavioral reasoning queries such as \textit{“Have I been skipping lunch more often recently?”} (see Figure~\ref{fig:teaser_fig}), which require tracking and comparing interactions distributed across multiple days rather than retrieving a single relevant event. A possible solution is to record interactions explicitly as structured records without summarization. However, this approach introduces a scalability challenge, as wearable cameras generate thousands of observations per day many of which correspond to repeated interactions.  Consequently, three requirements for effective long-horizon egocentric memory emerge: preserving interaction structure, supporting temporal aggregation across recurring interactions, and remaining scalable as observations accumulate. 

We address these requirements through \textbf{Imprint}, a structured memory framework in which each interaction is represented as an Interaction Record containing the interacting person, action, object, and the start and end timestamps of the interaction. While this representation preserves the interaction semantics required for long-horizon reasoning, storing every Interaction Record is impractical even for a day (e.g. >10,000 interactions in 8 hours~\cite{yang2025egolife}) with rapidly growing and highly redundant memories. To address this challenge, we introduce an online compression approach where memory compression occurs as interaction records arrive preventing the need to store all the interaction records. Incoming interaction records are organized into recurring event patterns, an importance score for each interaction record is assigned to determine its priority, and important interaction records are consolidated into compact long-term memories. The importance score is designed using the principles of frequency, redundancy and distinctiveness inspired by human long-term memory consolidation~\cite{mcclelland1995there}. 
The consolidated memory is able to long-horizon behavioral query such as "Have I been skipping lunch more often recently?" with evidence as shown in Figure~\ref{fig:teaser_fig}.

We evaluate Imprint on EgoLifeQA~\cite{yang2025egolife}, an egocentric benchmark with  seven-day of videos containing questions that require reasoning over interactions occurring from minutes to multiple days before the query. Compared with hierarchical caption summaries, Imprint achieves higher question answering accuracy while increasing evidence-grounded answers by $6\times$. This suggests that cognitively inspired importance signals improve long-horizon memory by preferentially retaining interactions that are most likely to support future retrieval and reasoning. Furthermore, we show that Imprint substantially reduces memory footprint and retrieval latency, demonstrating that interaction-centric memory compression is a scalable alternative to hierarchical text summarization for long-horizon egocentric reasoning. Our contributions are as follows:
\begin{itemize}

\item We introduce Imprint, a cognitively inspired online memory compression framework that discovers recurring interaction patterns, estimates their long-term importance, and selectively consolidates redundant interactions while preserving retrieval-relevant evidence.

\item Imprint showing that it retains relevant interactions better than hierarchical caption summarization, yielding higher QA accuracy, and $6\times$ better evidence grounding with the same LLM. 

\item We also show that Imprint requires $2.3\times$ lower memory and is $11.8\times$ faster at answering than hierarchical caption summaries.

\end{itemize}
\section{Related Work}
\label{sec:literature}
\paragraph{Long-Horizon Egocentric QA.} Early egocentric datasets such as CharadesEgo~\cite{sigurdsson2018charades}, EGTEA Gaze+~\cite{li2018eye}, EPIC-KITCHENS~\cite{damen2020epic}, and Ego4D~\cite{grauman2022ego4d} focus on activities ranging from seconds to a few hours, but do not capture the multi-day interaction dynamics required for persistent memory tasks. Recent work has begun addressing this limitation. EgoLife~\cite{yang2025egolife} introduces a 300-hour multimodal egocentric dataset together with the EgoLifeQA benchmark spanning five reasoning categories across seven days of continuous recording. 
Castle~\cite{rosetto2024castle} introduces a 600-hour multimodal egocentric dataset recorded over four days and includes the point of view of 10 participants.
Recent instruction-tuned multimodal LLMs ~\cite{zhang2023video,maaz2023video,li2024llava}, have improved video-language alignment and grounding. Retrieval-augmented approaches store past observations as captions or summaries and retrieve relevant entries during inference. EgoRAG~\cite{yang2025egolife} follows this paradigm by generating per-clip captions and organizing them hierarchically (clip $\rightarrow$ group $\rightarrow$ day). 
\paragraph{Memory-Augmented Retrieval for Long-Context Reasoning.} Recent work in long-context RAG highlights the limitations of unstructured text retrieval. Retrieval quality depends on preserving document structure and contextual continuity~\cite{laitenberger2025stronger}, while chunk-level retrieval often breaks semantic dependencies between related evidence~\cite{tao2025saki}. To improve scalability and coherence, recent systems incorporate graph-based indexing with dual-level retrieval~\cite{guo2024lightrag}, hierarchical coarse-to-fine abstraction~\cite{huang2025retrieval}, and external structured memory for persistent agents~\cite{yang2025efficient}. Together, these advances highlight the importance of structured memory representations for scalable long-horizon retrieval and reasoning.
\paragraph{Cognitive Memory Models.}
Cognitive memory theory suggests that scalable retrieval emerges from selective and structured representations rather than uniform storage, with consolidation influenced by repetition, recency, distinctiveness, and salience~\cite{tulving1973encoding}.
Repeated interactions are gradually compressed into structured semantic representations, enabling efficient long-term retention while preserving salient contextual information~\cite{shaham2022stochastic, mcclelland1995there}.
At the same time, distinctive or behaviorally significant events tend to resist abstraction and remain individually retrievable in episodic memory~\cite{hunt2006distinctiveness}.
However, existing egocentric QA pipelines largely treat observations uniformly, overlooking principles of selective consolidation and salience-driven retention. In contrast, our framework models interactions as structured action–object memories with importance-aware weighting, enabling selective compression while preserving distinctive events for long-horizon retrieval.
\section{Imprint: Interaction-centric Memory Framework}
\subsection{Interaction Records}
\label{fio_rep}
Free-form captions often obscure the interaction structure required
for long-horizon reasoning. For example, repeated observations such
as \textit{pick up mug}, \textit{fill mug with coffee}, and
\textit{drink coffee} may be summarized as a generic description
such as \textit{having coffee}. While sufficient for describing an
individual event, such summaries make it difficult to track
recurring interactions, aggregate behavior over time, and answer
questions about habits and routines. To preserve interaction-level
semantics, we represent each observation as a structured interaction record:
\begin{equation}
f_i =
(p_i, a_i, o_i, t_i^{s}, t_i^{e})
\label{eq:fio}
\end{equation}
where $p_i$ denotes the interacting person, $a_i$ the action, 
$o_i$ the interacted object, $t_i^{s}$ and $t_i^{e}$ the interaction start and end timestamps. 
We parse each caption into an interaction record using an LLM ( \textsc{Qwen2.5-7B-Instruct})  as shown in Figure~\ref{fig:pipeline}. 
The extraction prompt and schema are provided in
Supplementary B.
\begin{figure*}[!htbp]
  \centering
  \includegraphics[width=\textwidth,]{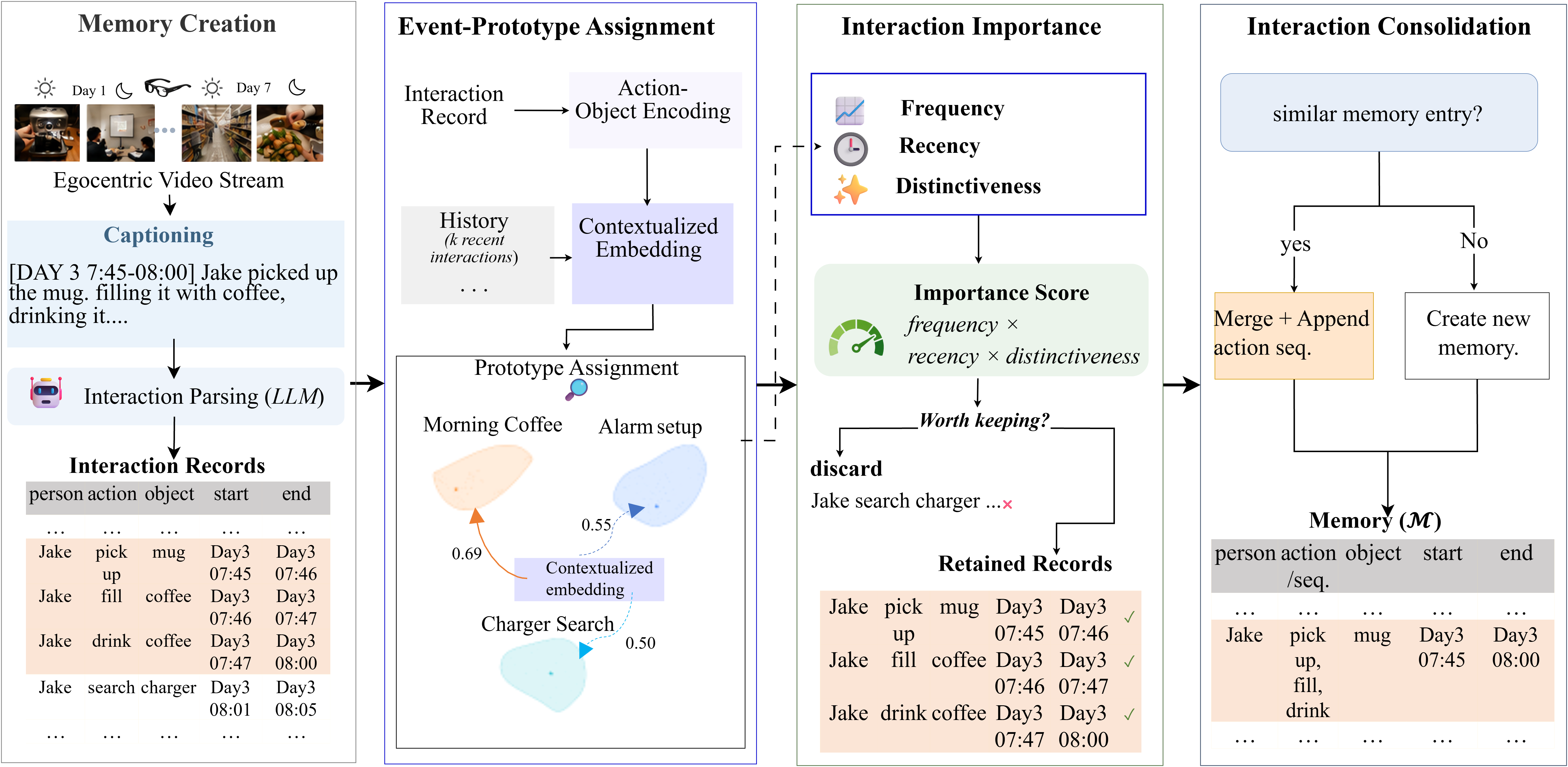}
  \caption{ Imprint online memory compression. From every egocentric video caption, an LLM produces a structured interaction record. Compression runs in three stages: each record is grouped into the nearest event prototype (Event-Prototype Assignment), filtered by an importance score (Interaction Importance), and merged with similar observations into the compact memory$
  \mathcal{M}$ (Interaction Consolidation)}
  \label{fig:pipeline}
\end{figure*}

\subsection{Memory Compression}
Unlike offline compression methods that operate after all observations have been collected, Imprint performs memory consolidation online for every incoming interaction record obtained from the caption. This streaming design eliminates the need to retain the complete set of extracted interaction records, allowing memory to evolve continuously as new observations arrive. 

\subsubsection{Event Prototype Assignment}
Interaction records generated from continuous egocentric observations often exhibit recurring patterns due to routine human behavior. For example, a user may repeatedly perform interactions such as \textit{pick up mug}, \textit{fill mug with coffee}, and \textit{drink coffee} every morning. Treating each occurrence as an independent memory introduces substantial redundancy while obscuring recurring behavioral patterns. To identify such patterns and estimate their long-term significance, we organize incoming interaction records into a set of event prototypes $\mathcal{G}$. Each event prototype $g_n \in \mathcal{G}$ maintains a recurrence count $\nu_n$ and last-seen timestamp $T_n^{\mathrm{last}}$. These statistics are later used to compute frequency and recency signals for interaction importance estimation.

For each incoming record \(f_t\), we first encode its action and object as \( e_t=\mathrm{Encoder}(a_t,o_t)\) and
augment it with short-term interaction history of $k$ recent interactions to form a contextualized embedding \(\tilde e_t\).
The contextualized embedding of the interaction record is compared against all prototypes in $\mathcal{G}$ and assigned to the nearest prototype  subject to the similarity exceeding $\delta_g$. Otherwise, a new prototype is created to represent the unseen interaction pattern. To maintain bounded memory growth, the prototype bank is capped at $N_{\max}$. Once this capacity is reached, incoming interactions are assigned to their nearest existing prototype, trading representational granularity for scalability.
Event prototypes are auxiliary structures rather than retrievable memories - they summarize recurring interaction patterns and maintain the recurrence and recency statistics used for downstream importance estimation. Implementation details are provided in Supplementary C.

\subsubsection{Interaction Importance}
Not all recurring interactions contribute equally to long-term memory. For instance, a coffee-drinking interaction observed every morning over several days is likely to be more informative for future behavioral reasoning than an isolated interaction such as searching for a receipt once. Inspired by cognitive theories of memory consolidation~\cite{shaham2022stochastic,mcclelland1995there,hunt2006distinctiveness}, we prioritize retrieval-relevant interactions by assigning an importance score to each interaction record based on cognitive concepts of frequency, recency and diversity. The importance score for an interaction record $f_t$ is computed as follows:

\begin{equation}
w(f_t)
=
\overbrace{\log(1+ \nu_{n_t})}^{\text{\small frequency}} \cdot \overbrace{
e^{-\lambda \Delta T}}^{\text{\small recency}}
{\overbrace{\frac{1}{1+\log(1+N_t)}}^{\text{\small distinctiveness}}}
\end{equation}
where $n_t$ denotes the event prototype assigned to the interaction record $f_t$, $\nu_{n_t}$ denotes the recurrence count of the event prototype $g_{n_t}$. $\Delta T$ shows the elapsed time since its previous occurrence, and $\frac{1}{1+\log(1+N_t)}$ reflects the diversity of event prototypes ($N_t$) currently in memory. Distinctiveness is significantly higher for the initial interaction records and gradually decreases when the prototypes are populated.

The importance score $w(f_t)$ estimates the long-term relevance of an interaction by combining recurrence and recency signals and $\lambda$ controls the rate at which importance decays with temporal distance. Records with $w(f_t)$ greater than a threshold ($\tau$) are retained. In addition to the retention decision, the importance score is recorded along with the retained interaction record, and then reused for Retrieval and Question Answering (Section~\ref{sec:r_QA}).

\subsubsection{Interaction Consolidation}
Even after importance filtering, we may have multiple instances of the same interaction pattern across days. For example, repeated records corresponding to \textit{drink coffee} observed across several mornings would be stored independently despite describing essentially the same recurring behavior. Storing such interactions separately increases memory redundancy and retrieval cost.

For each retained interaction record $f_i$ from the previous stage, we compare its action $a_i$, object $o_i$, and contextual embedding $\tilde e_i$ against existing entries in the compressed memory $\mathcal{M}$. An interaction record is merged with an existing memory entry when their action-object representations and contextual embeddings are similar. For example, interactions such as \textit{open book}, \textit{read notes}, and \textit{write on paper} may be consolidated into a single event sequence representing a studying episode. When no suitable match exists, a new memory entry is created for the interaction record and added to $\mathcal{M}$.
For instance, the consolidated \textit{studying} interactions above form a single
entry
$m = (\langle f_1, f_2, f_3 \rangle,\, w)$, where
$f_1 = (\textit{Jake},\textit{open},\textit{book},t^s_1,t^e_1)$ and similarly, $f_2, f_3$ are \textit{read notes} and \textit{write on paper}
records, and $w$ is their aggregated importance.
Each memory entry preserves the temporal ordering, aggregated importance, and contextual evidence of the interactions it represents, enabling reconstruction of events during retrieval. 

\subsection{Retrieval and Question Answering}
\label{sec:r_QA}
Given a question $q$ asked at time $t_q$, interaction records whose start time precedes the question timestamp are obtained \( \mathcal{M}_q = \{m_k \in \mathcal{M}
\mid t_k^{s} \le t_q\}\). The question is then analyzed using an LLM to extract retrieval cues, including object $o_q$, action $a_q$, and temporal constraints $r_q$ (if present). For example, for the question \textit{``When did Jake cook pasta in the evening?''}, the extracted retrieval cues are $o_q=\textit{pasta}$, $a_q=\textit{cook}$, and $r_q=\textit{evening}$. The extracted object and action are encoded using the same sentence encoder employed during memory construction. Each candidate memory entry $m_k$ is then ranked according to a weighted combination of object and action similarity:

\begin{equation}
s_k =
\gamma_1 \cos(e_{o_q}, e_{o_k})
+
\gamma_2 \cos(e_{a_q}, e_{a_k})
+\gamma_3 w(f_k),
\label{eq:3}
\end{equation}

where $e_{o_k}$ and $e_{a_k}$ denote the object and action embeddings and $w_k$ importance score associated with the memory entry $m_k$. The top-$k$ highest-scoring memories are retrieved as evidence 
\[ R = \mathrm{TopK}
\bigl(\{(m_k,s_k)\} \forall {m_k \in \mathcal{M}_q}\bigr).
\]
For questions containing temporal constraints, the retrieved evidence is further filtered according to $r_q$. Specifically, \texttt{first\_time} and \texttt{last\_time} select the earliest and latest matching interactions, respectively, while \texttt{before} and \texttt{after} retain only interactions satisfying the corresponding temporal condition. The resulting evidence set $R'$ is then provided to an LLM for answer generation along with the question: \(r = \mathrm{LLM}(q,R') \). More details about the retrieval mechanism are in Supplementary D.
\section{Experiments and Results}
\label{sec:experiments}
\subsection{Setup}
\label{sec:setup}
\paragraph{Dataset.}
EgoLifeQA~\cite{yang2025egolife} is a long-horizon egocentric QA benchmark of 3,000 multiple-choice questions, derived from seven consecutive days of wearable-camera recordings of multiple participants. We evaluate on the published 500-question benchmark for the participant Jake. Questions are distributed across five reasoning categories-\textsc{EntityLog} (125), \textsc{EventRecall} (126), \textsc{RelationMap} (125), \textsc{HabitInsight} (61), and \textsc{TaskMaster} (63). 
\paragraph{Implementation details.}
We compare Imprint against EgoRAG~\cite{yang2025egolife}, a hierarchical caption-retrieval framework that uses GPT-4o as the answering LLM. To isolate the contribution of memory representation from LLM capacity, we reproduce EgoRAG~\cite{yang2025egolife} with \textsc{Qwen2.5-7B-Instruct}, the same LLM used throughout our pipeline. This reproduced version serves as our primary baseline. 
We use \textsc{EgoGPT}~\cite{yang2025egolife} for caption generation, identical to the captioner used in EgoRAG~\cite{yang2025egolife}, ensuring that performance differences arise from memory representation and retrieval design rather than caption quality. We use a query-aware prompting strategy for caption generation. Details are in the Supplementary A. For all subsequent stages, including extraction of interaction record, compression and answer generation, we use \textsc{Qwen2.5-7B-Instruct}.
For event prototype assignment, each interaction record is augmented with a short-term history of the previous $k=5$ interactions to form the contextualized embedding $\tilde e_t$. Historical interactions are incorporated using a context-history weight of $\alpha_h=0.5$, and only interactions whose similarity exceeds $\delta_h=0.3$ contribute to the contextual representation. Incoming interaction records are assigned to the nearest event prototype when cosine similarity exceeds $\delta_g=0.75$; otherwise, a new prototype is created. The prototype bank is capped at $300$ entries. 

For Interaction Importance, the recency decay coefficient is set to $\lambda =1.14$ and retain interaction records whose importance score satisfies $w(f_i) \geq \tau$, where $\tau = 0.2$. During Interaction Consolidation, retained records are merged into existing memory entries when their action-object representations and contextual embeddings exceed a cosine similarity threshold of $\delta_m = 0.75$. 

At query time, candidate memory entries are retrieved from the compressed memory $\mathcal{M}$ and ranked using the retrieval score defined in Eq.~\ref{eq:3}, with weights $\gamma_1 = 0.53$, $\gamma_2 = 0.29$, and $\gamma_3 = 0.18$ details are in Supplementary E.4. The $TopK=5$ memories are provided to \textsc{Qwen2.5-7B-Instruct} for answer generation. All experiments are conducted on a single NVIDIA L40 GPU.

\paragraph{Metrics.}
For long-horizon egocentric assistants, accuracy alone is insufficient because language models can answer plausibly using parametric priors or semantic guessing without retrieving supporting memory evidence. This is particularly problematic for personal reasoning queries such as \textit{“Do my recent activities suggest changes in my health?”}, which require aggregating temporally distributed interactions rather than relying on generic behavioral assumptions. To evaluate whether answers are genuinely grounded in retrieved memory, we introduce grounded accuracy in addition to QA accuracy.

\textbf{Grounded Accuracy (GA)} measures the proportion of correct predictions supported by retrieved evidence rather than inferred solely from the model's parametric knowledge:

\[
\mathrm{GA} =
\frac{
\#\ \text{grounded correct answers}
}{
\#\ \text{correct answers}
}
\]
An answer is grounded when it is obtained directly from a retrieved
interaction and its explanation matches that interaction's fields. For the question
\textit{What happened the last time I was at the claw machine?}, selecting
the correct option \textit{Distribute coins} is grounded when its
explanation matches a retrieved record, e.g.\
$\langle \textit{Jake}, \textit{operated}, \textit{claw machine}, \textit{coins} \rangle$,
whose \textit{coins} field matches the answer. The same correct option is not grounded when its explanation matches no retrieved field and instead
relies on a general assumption (e.g.\ that arcades involve coins) but not evidence-supported. 
More examples are in Supplementary E.5.

\subsection{Interaction-Centric Retrieval Performance}
Table~\ref{tab:main_results} reveals that the gains of Imprint arise from both from the use of structured interaction representation and memory compression. Replacing hierarchical caption summaries in EgoRAG~\cite{yang2025egolife} with Interaction Records dramatically improves grounding (GA: 10.8\% $\rightarrow$ 41.5\%), indicating that explicit person-action-object representations allow better evidence retrieval than textual summaries. However, this improvement does not translate directly into higher QA accuracy, which slightly decreases from 31.0\% to 28.4\%. 
We hypothesize that while raw Interaction Records preserve relevant evidence, they also introduce substantial redundancy, retrieval noise, and competition among repeated observations of the same behavior. Imprint's memory compression improves both grounding and QA accuracy, demonstrating that effective long-horizon memory requires not only preserving interaction structure but also organizing that structure into a retrieval-efficient representation.
\begin{table}[!htbp]
\centering
\small
\setlength{\tabcolsep}{2pt}

\begin{tabular}{lccc}
\toprule
\textbf{Method} &
\textbf{Acc.} &
\textbf{GA} \\
\midrule

\makecell[l]{\textcolor{gray}{EgoRAG~\cite{yang2025egolife}}
\textcolor{gray}{(GPT-4o)}}
&
\textcolor{gray}{36.0}
&
\textcolor{gray}{--}
\\

\makecell[l]{EgoRAG~\cite{yang2025egolife}(Qwen2.5-7B)}
&
31.0 
&10.8
\\
All Interaction Records (Qwen2.5-7B)
&
28.4 &
41.5
\\
Imprint (Qwen2.5-7B)
&\textbf{35.8} 
&\textbf{64.8}
\\
\bottomrule
\end{tabular}
\caption{
QA performance across memory frameworks.
}
\label{tab:main_results}
\end{table}
\begin{figure}[!htbp]
  \centering
  \includegraphics[width=\columnwidth]{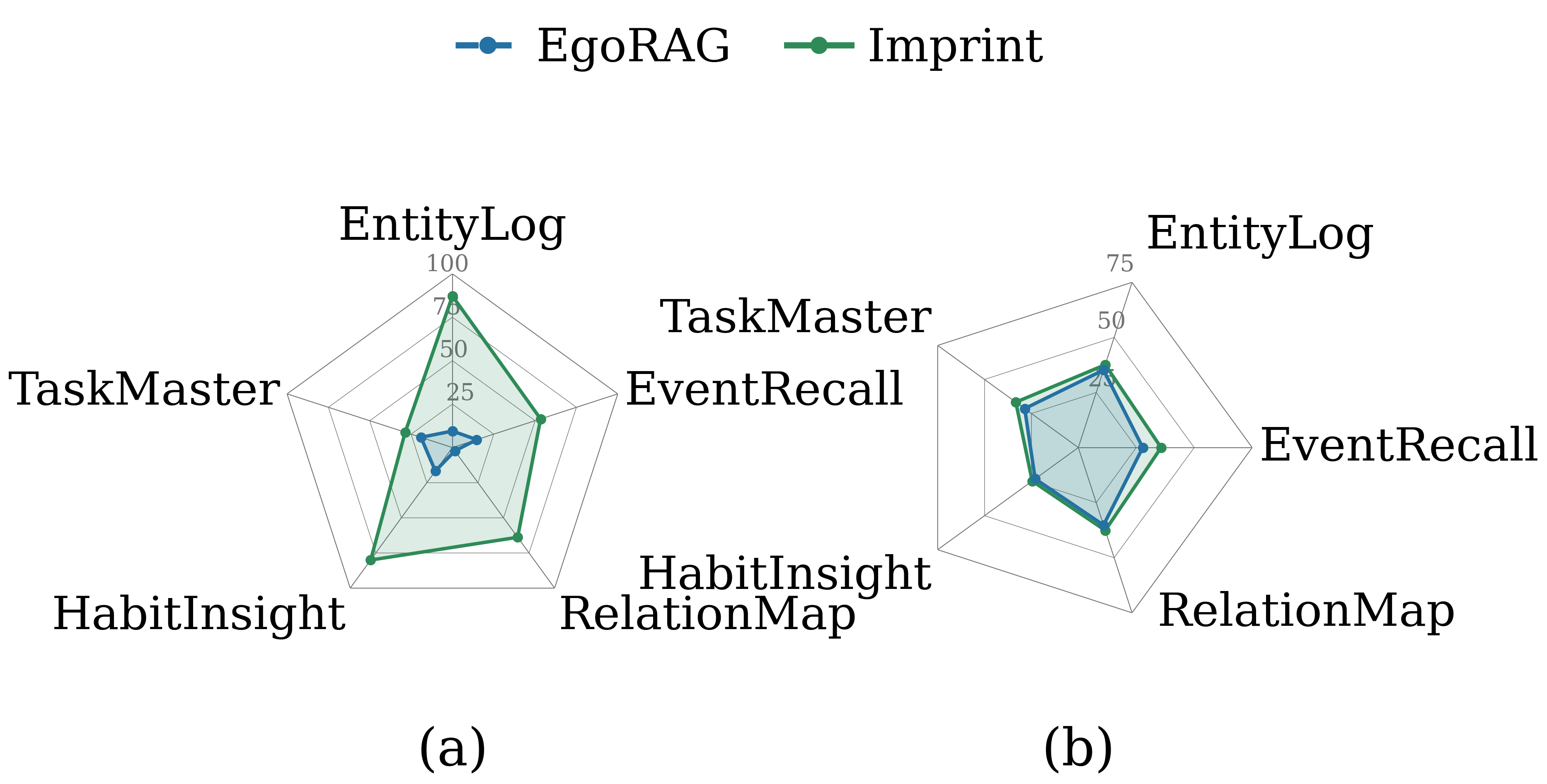}
  \caption{EgoLifeQA~\cite{yang2025egolife} category-wise comparison of Imprint and EgoRAG (Qwen2.5-7B-Instruct) (a) Grounded Accuracy (GA\%) and (b) QA Accuracy. Exact numerical values for each category are in Supplementary E.1.}
  \label{fig:radar_plot}
\end{figure}
Figure~\ref{fig:radar_plot} provides a category-wise analysis across the five EgoLifeQA question types. Imprint improves grounded answering across all categories, with the largest gains observed on EntityLog, EventRecall, RelationMap, and HabitInsight. These categories require tracking entities across time, recovering relationships between people, actions, and objects, and aggregating evidence across multiple interactions. 
Such reasoning benefits directly from interaction records, where interaction participants and objects are represented explicitly rather than embedded within natural-language summaries. Improvements on HabitInsight further suggest that Imprint preserves recurring interaction patterns that are critical for habit-level reasoning. In contrast, the gains on TaskMaster are comparatively smaller, indicating that they need planning and multi-step reasoning, so better memory alone is not enough to improve performance. Collectively, these results support the central hypothesis of Imprint: long-horizon egocentric QA benefits from memories that preserve interaction structure and consolidate interaction records around behaviorally salient recurring interactions.

\subsection{Long-Horizon Retrieval and Efficiency}
\begin{figure}[!htbp]
  \centering
\includegraphics[width=\columnwidth, trim=0 0 0 0,  clip]{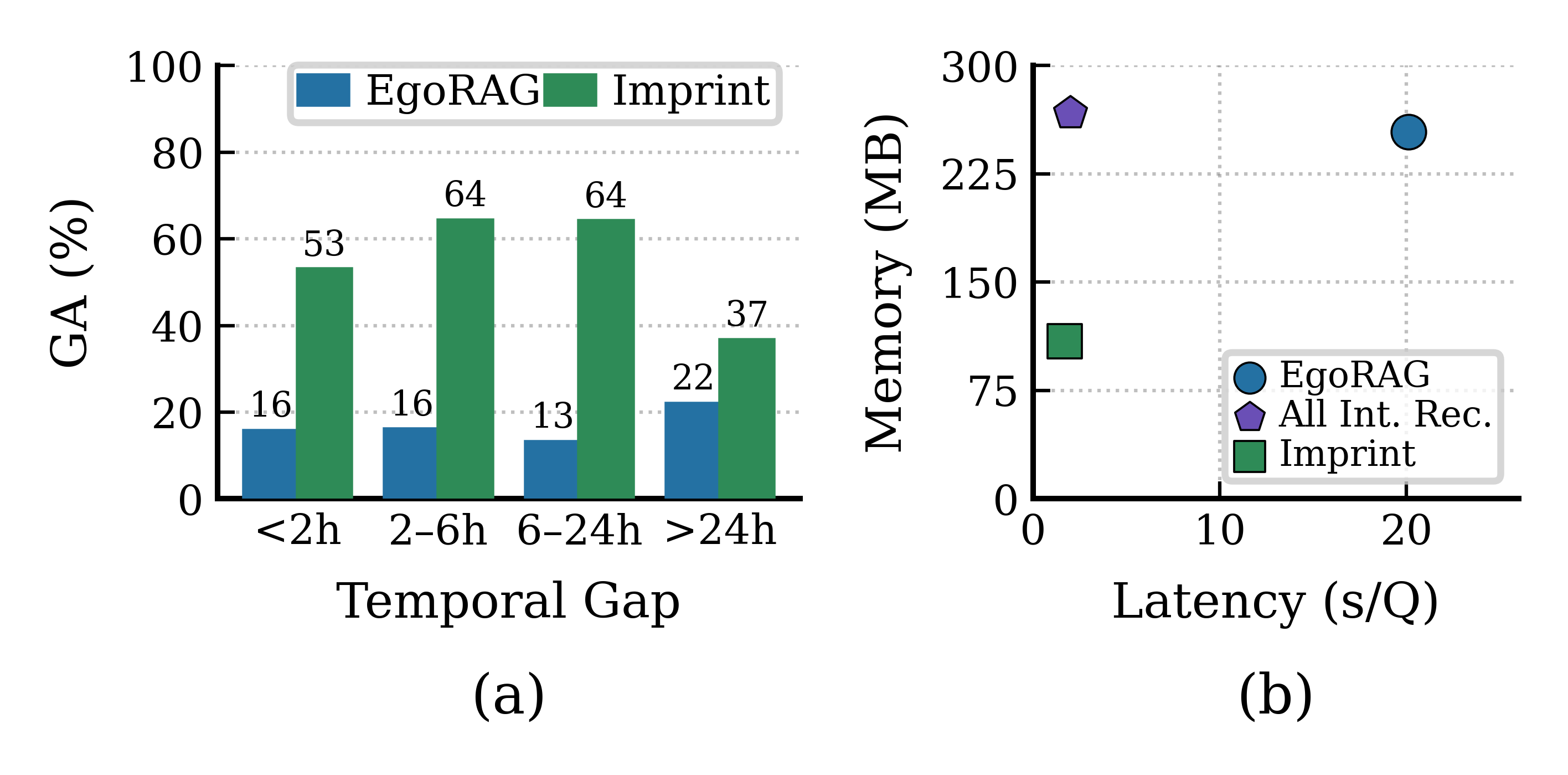}
  \caption{Comparison of retrieval efficiency. (a) Temporal gaps (query time to event time), and 
  (b) Memory vs.\ retrieval latency.}
  \label{fig:tempo_mem_combined}
\end{figure}

Figure~\ref{fig:tempo_mem_combined}(a) evaluates retrieval grounding using the temporal-gap partitions introduced by EgoRAG~\cite{yang2025egolife}, which group questions according to the time elapsed between a query and its supporting evidence. Across all temporal gaps, Imprint substantially outperforms EgoRAG in Grounded Answers (GA). Notably, Imprint's grounded accuracy remains between 53-64\% for evidence occurring up to 24 hours before the query, whereas EgoRAG remains below 22\% across all gaps. The performance drop in the >24h gap likely reflects the increased difficulty of retrieving and aggregating evidence distributed across multiple days of experience. Nevertheless, Imprint retains a clear advantage over EgoRAG, indicating that interaction-centric memories remain more robust to temporal separation than summary-based representations.

Figure~\ref{fig:tempo_mem_combined}(b) examines the efficiency implications of memory consolidation. Despite achieving substantially higher retrieval grounding, Imprint uses less memory and retrieves evidence significantly faster than both All Interaction Records1(All Int. Rec.) and EgoRAG. Memory size decreases from 267 MB for all interaction records to 109 MB after consolidation, compared with EgoRAG's 254.2 MB consisting of both caption-level memories and hierarchical summary. More importantly, retrieval latency decreases from 20.1 s/query in EgoRAG to 1.7 s/query in Imprint, yielding an 11.8$\times$ speedup. 
These results indicate that Imprint simultaneously improves retrieval grounding while reducing storage and retrieval cost, making it well-suited for long-horizon egocentric memory, despite continuously growing recordings.

\subsection{Ablations}
\paragraph{Stage-wise Memory Compression.}
Table~\ref{tab:stage_ablation} evaluates the contribution of each stage in the Imprint memory formation pipeline. Removing any stage reduces both Grounded Answers (GA) and QA Accuracy, indicating that event-prototype assignment, interaction importance, and interaction consolidation each contribute to effective long-horizon reasoning. The largest reduction in GA occurs when event-prototype assignment is removed (64.8\% $\rightarrow$ 58.6\%), accompanied by a notable drop in accuracy (35.8\% $\rightarrow$ 31.4\%). This suggests that identifying recurring interaction patterns is critical for estimating memory relevance and distinguishing routine behaviors from isolated observations. Removing interaction importance yields the largest accuracy degradation after prototype assignment (35.8\% $\rightarrow$ 34.2\%) while reducing GA to 57.3\%, indicating that prioritizing interactions according to recurrence, recency, and distinctiveness improves both evidence selection and answer quality. Similarly, removing interaction consolidation decreases GA to 58.3\% and accuracy to 32.6\%, demonstrating that preserving interaction structure alone is insufficient; repeated observations must also be organized into compact memory entries to reduce redundancy and improve retrieval efficiency. The final row further shows that interaction consolidation alone is insufficient. Notably, the relative variation across configurations is larger for GA than for accuracy, suggesting that these stages primarily improve the quality of retrieved evidence, with improved QA accuracy emerging as a consequence of better-grounded retrieval. 
\begin{table}[!htbp]
\centering
\scriptsize
\begin{tabular}{lcc}
\toprule
\textbf{Configuration}
& \textbf{GA/Acc.} \\
\midrule
w/o Event-Prototype Assignment
& 58.60 / 31.4 \\
w/o Interaction Importance
 & 57.31 / 34.2 \\
w/o Interaction Consolidation
& 58.28 / 32.6 \\
Interaction Consolidation only
& 56.17 / 32.4 \\
\textbf{Imprint }
& \textbf{64.80} / \textbf{35.80} \\
\bottomrule
\end{tabular}
\caption{Stage-wise ablation of the memory compression.}
\label{tab:stage_ablation}
\end{table}
\paragraph{Interaction Importance Signals.} 
Table~\ref{tab:cognitive_ablation} reveals a consistent trend across both Grounded Accuracy (GA) and Accuracy: removing any component of the importance score degrades performance, indicating that frequency, recency, and distinctiveness each contribute useful information for memory selection. 
Removing frequency produces the largest performance degradation indicating that recurrence is the strongest signal for identifying retrieval-relevant interactions. 
Removing recency also substantially reduces performance, reflecting the fact that many questions emphasize recent interactions rather than equally frequent events from the distant past. 
Distinctiveness has the smallest impact, but its removal still degrades both GA and Accuracy, indicating its role in distinguishing similar interaction patterns. Notably, the effect of all three signals is substantially larger on GA than on accuracy, indicating that their primary benefit is improving retrieval quality and evidence grounding. These findings support the central premise of Imprint that it is able to selectively retain behaviorally relevant interactions.
\begin{table}[!htbp]
\centering
\scriptsize
\begin{tabular}{cccc}
\toprule
\textbf{Freq.} & \textbf{Rec.} & \textbf{Dist.}  & \textbf{GA/Acc.}  \\
\midrule
$\times$
& $\checkmark$
& $\checkmark$
& 38.19/28.00
 \\
$\checkmark$
& $\times$
& $\checkmark$
& 51.32/28.20
 \\
$\checkmark$
& $\checkmark$
& $\times$
& 54.79/29.20 
 \\
$\checkmark$
& $\checkmark$
& $\checkmark$
& \textbf{64.80}/\textbf{35.80}
 \\
\bottomrule
\end{tabular}
\caption{Interaction Importance ablation on EgoLifeQA. Entries are GA/Acc. \% ; $\Delta$ denotes reduction relative to Imprint.}
\label{tab:cognitive_ablation}
\end{table}
\paragraph{Memory Growth and Prototype Behavior.}
\begin{figure}[!htbp]
  \centering
\includegraphics[width=\columnwidth,trim=0 0 0 0, clip ]{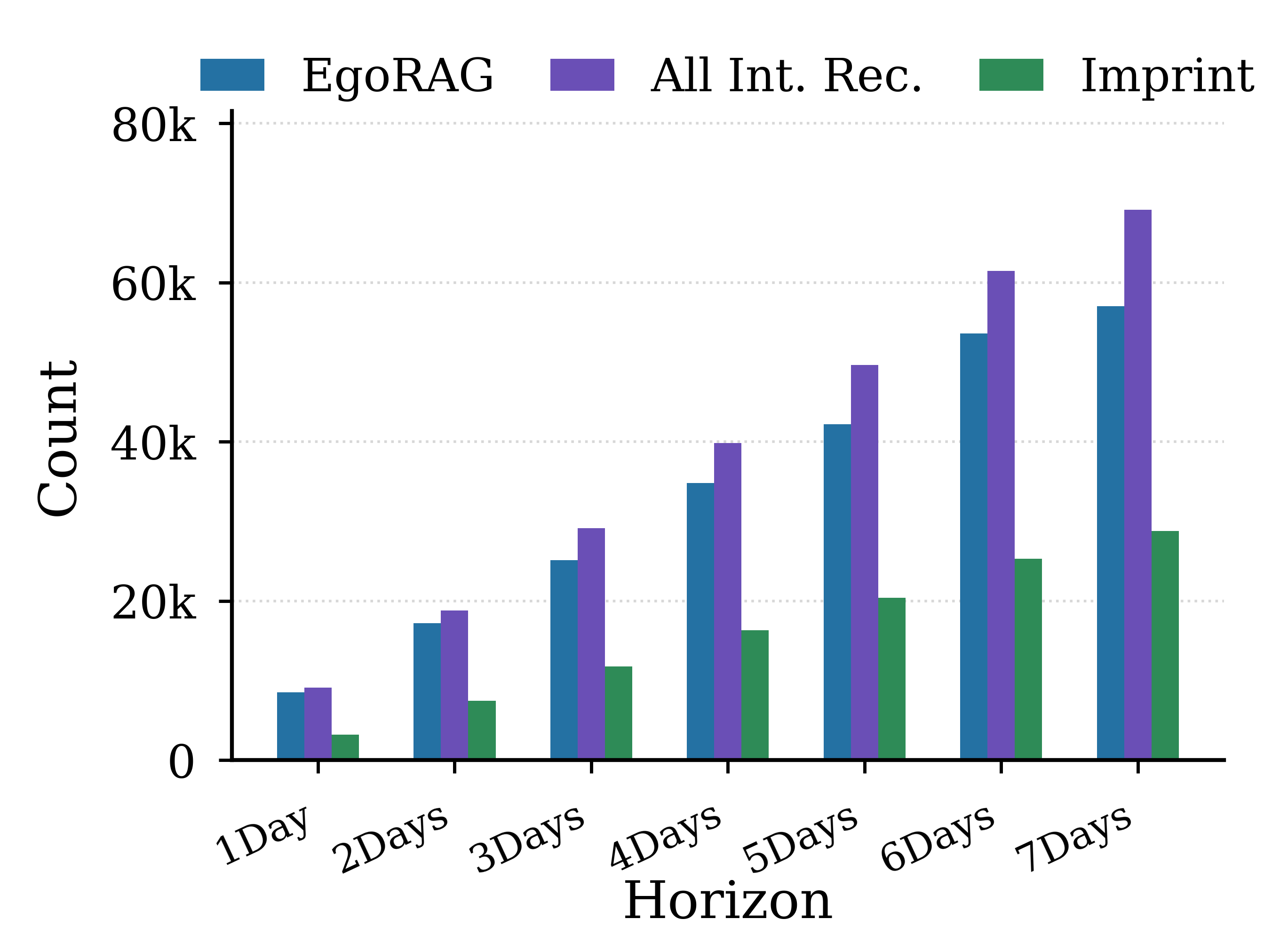}
\caption{Growth of memory over seven days for Imprint vs. EgoRAG}
\label{fig:growth_irm}
\end{figure}

Figure~\ref{fig:growth_irm} illustrates the growth of memory over the seven-day EgoLife~\cite{yang2025egolife} recording. Imprint grows substantially more slowly than EgoRAG's summaries (day, hour, minute) as recording length increases. The gap widens after Day 3, when recurring interaction patterns become increasingly common and are consolidated into existing memories rather than stored as independent entries. By Day 7, Imprint maintains approximately 2.4$\times$ fewer interaction records in memory as seen in its comparison with all interaction records. Importantly, this improved scalability does not incur significant pre-processing overhead: compressing the full seven-day EgoLife recording requires 3 h 4 min, comparable to EgoRAG's 2 h 54 min hierarchical summarization pipeline.

\begin{figure}[!htbp]
  \centering
  \includegraphics[width=\columnwidth,trim={0 0 0 0},clip]{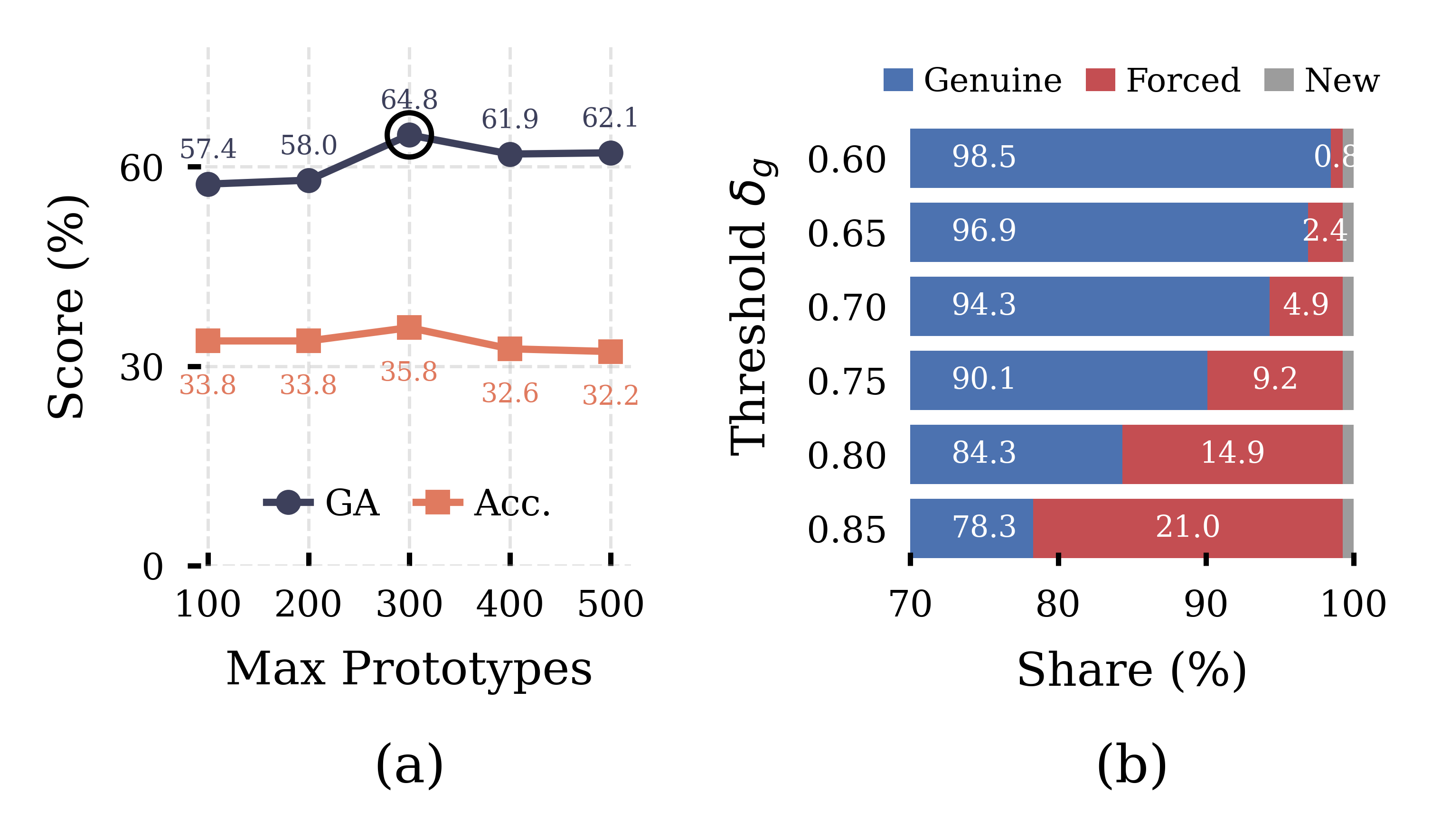}
  \caption{Event Prototype assignment behavior under different similarity thresholds ($\delta_g$). (a) GA and Accuracy vs. prototype-cap size. (b) For each threshold $\delta_g$, the fraction of interactions resulting in genuine merges (similarity above $\delta_g$), forced merges ($N_{\max}$reached), and new prototypes created when similarity is below $\delta_g$.}
  \label{fig:prototype_assignment_composition}
\end{figure}

Figure~\ref{fig:prototype_assignment_composition} provides additional insight into the consolidation mechanism. As shown in Figure~\ref{fig:prototype_assignment_composition}(a), too few prototypes (100-200) force semantically distinct interaction patterns share prototypes, reducing the quality of importance estimation and subsequent consolidation. Increasing the budget beyond 300 provides little additional benefit and slightly degrades performance, suggesting that excessive prototype fragmentation weakens the recurrence statistics needed for reliable memory consolidation. Figure~\ref{fig:prototype_assignment_composition}(b) shows that at event prototype merging threshold $\delta_g=0.75$, more than 90\% of incoming interactions are assigned to existing prototypes, while only a small fraction create new prototypes. This suggests that most everyday observations correspond to recurring behavioral patterns rather than novel interaction types. 

\begin{figure}[!htbp]
  \centering
\includegraphics[width=\columnwidth, trim=0 0 0 0, clip ]{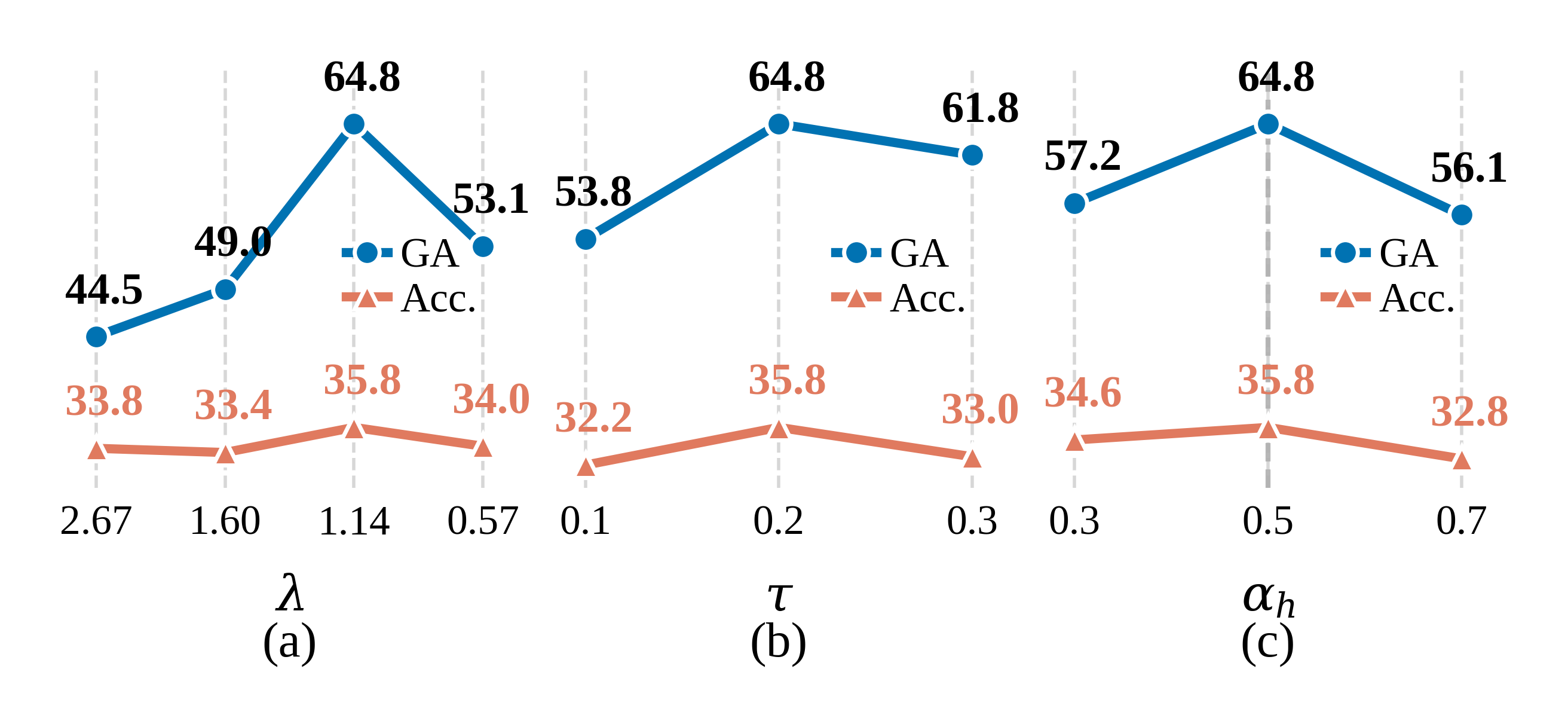}
\caption{Hyperparameter ablations. (a) Impact of  $\lambda$ which controls the rate at which importance
decays with temporal distance. (b)  Impact of threshold $\tau$ on $w(f_t)$.
(c) Impact of context-history weight $\alpha_h$.}
  \label{fig:key_hyperparameter}
\end{figure}

\paragraph{Impact of Hyperparameters.} We study the effect of hyperparameters on GA (see Figure~\ref{fig:key_hyperparameter}). As shown in Figure ~\ref{fig:key_hyperparameter}(a) at  $\lambda=1.14$ gives the highest GA.
We observe that a higher value of $\lambda$ suppresses interactions that remain relevant over longer horizons, while a lesser value of $\lambda$ reduces the ability to prioritize recent evidence.  Next, we show the effect of importance score threshold $\tau$ on the retention of records, as shown in Figure~\ref{fig:key_hyperparameter}(b). The performance at $\tau=0.2$ strikes a balance between retaining interaction records and limiting redundant interactions. Finally, we study the impact of $\alpha_h$ in Figure~\ref{fig:key_hyperparameter}(c). We find moderate $\alpha_h=0.5$ produces the best result, while both weaker and stronger history weighting reduce retrieval quality. Thus, over-weighting historical context biases prototype assignment towards outdated observations, degrading adaptation to evolving egocentric video streams. 
Other hyperparameter ablations on the history length $k$ and history-similarity threshold $\delta_h$ are provided in Supplementary Section E.3.

\begin{figure}[!htbp]
  \centering
  \includegraphics[width=\columnwidth]{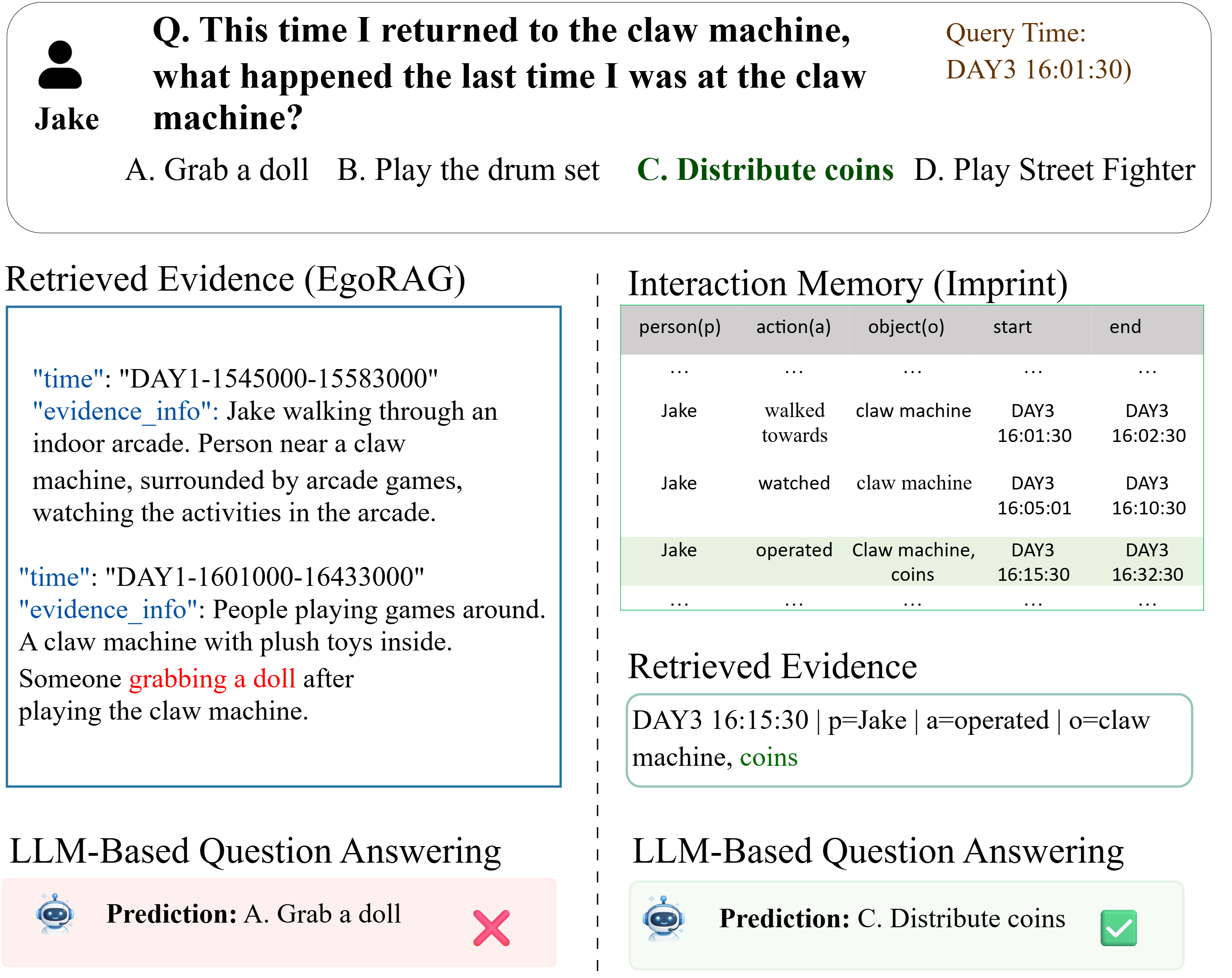}
  \caption{Comparison of retrieved evidence for an event recall query. 
Imprint retrieves the relevant interaction (\textit{Jake--operated--claw machine, coins}) at the target time, whereas EgoRAG fails to retrieve grounding evidence.}
  \label{fig:Qualitative}
\end{figure}
\paragraph{Qualitative Analysis} Figure~\ref{fig:Qualitative} shows that structured interaction memory improves intra-day episodic recall.
For the question \textit{``What happened the last time I was at the claw machine?''} (DAY3 16:01),
Imprint retrieves a temporally grounded interaction sequence from DAY3 16:01--16:32, preserving the progression \textit{walked towards $\rightarrow$ watched $\rightarrow$ operated} around the claw-machine area. Crucially, the retrieved interaction records retain discriminative interaction objects such as \textit{coins} and \textit{claw machine}, directly grounding the correct answer \textbf{``Distribute coins.''} In contrast, EgoRAG~\cite{yang2025egolife} retrieves semantically related but shallow location-centric captions (e.g., \textit{``standing near,'' ``visible nearby,'' ``looking at claw machines''}) and defaults to the prototypical arcade prior \textbf{``Grab a doll.''}
These examples highlight how interaction-centric memory better supports temporally grounded episodic retrieval in long-horizon egocentric QA.

\section{Conclusion}
\label{sec: Conclusion}
In this work, we introduced Imprint, a structured memory framework for long-horizon egocentric question answering. By consolidating recurring interactions using a cognitively inspired importance score, Imprint stores fewer interactions while producing more evidence-grounded answers. Experiments on EgoLifeQA~\cite{yang2025egolife} show that Imprint improves grounded answers while substantially reducing memory footprint and retrieval latency. However, its performance remains dependent on caption quality, as errors in identifying people, objects, or locations can propagate through memory formation and retrieval. While our importance score is based on cognitively motivated heuristics, future work could explore learnable memory heuristics for adaptive memory.
{
    \small
    \bibliographystyle{ieeenat_fullname}
    \bibliography{main}

@String(ECCV= {Eur. Conf. Comput. Vis.})

@String(ECCV  = {ECCV})

@inproceedings{laitenberger2025stronger,
  title={Stronger baselines for retrieval-augmented generation with long-context language models},
  author={Laitenberger, Alex and Manning, Christopher D and Liu, Nelson F},
  booktitle={Proceedings of the 2025 Conference on Empirical Methods in Natural Language Processing},
  pages={32547--32557},
  year={2025}
}

@inproceedings{tao2025saki,
  title={SAKI-RAG: Mitigating Context Fragmentation in Long-Document RAG via Sentence-level Attention Knowledge Integration},
  author={Tao, Wenyu and Xing, Xiaofen and Li, Zeliang and Xu, Xiangmin},
  booktitle={Proceedings of the 2025 Conference on Empirical Methods in Natural Language Processing},
  pages={1195--1213},
  year={2025}
}

@article{guo2024lightrag,
  title={Lightrag: Simple and fast retrieval-augmented generation},
  author={Guo, Zirui and Xia, Lianghao and Yu, Yanhua and Ao, Tian and Huang, Chao},
  journal={arXiv preprint arXiv:2410.05779},
  volume={2},
  number={3},
  year={2024}
}

@article{huang2025retrieval,
  title={Retrieval-augmented generation with hierarchical knowledge},
  author={Huang, Haoyu and Huang, Yongfeng and Yang, Junjie and Pan, Zhenyu and Chen, Yongqiang and Ma, Kaili and Chen, Hongzhi and Cheng, James},
  journal={arXiv preprint arXiv:2503.10150},
  year={2025}
}

@inproceedings{yang2025efficient,
  title={Efficient integration of external knowledge to LLM-based world models via retrieval-augmented generation and reinforcement learning},
  author={Yang, Chang and Wang, Xinrun and Zhang, Qinggang and Jiang, Qi and Huang, Xiao},
  booktitle={Findings of the Association for Computational Linguistics: EMNLP 2025},
  pages={9484--9501},
  year={2025}
}

@article{sigurdsson2018charades,
  title={Charades-ego: A large-scale dataset of paired third and first person videos},
  author={Sigurdsson, Gunnar A and Gupta, Abhinav and Schmid, Cordelia and Farhadi, Ali and Alahari, Karteek},
  journal={arXiv preprint arXiv:1804.09626},
  year={2018}
}

@inproceedings{li2018eye,
  title={In the eye of beholder: Joint learning of gaze and actions in first person video},
  author={Li, Yin and Liu, Miao and Rehg, James M},
  booktitle={Proceedings of the European conference on computer vision (ECCV)},
  pages={619--635},
  year={2018}
}

@article{damen2020epic,
  title={The epic-kitchens dataset: Collection, challenges and baselines},
  author={Damen, Dima and Doughty, Hazel and Farinella, Giovanni Maria and Fidler, Sanja and Furnari, Antonino and Kazakos, Evangelos and Moltisanti, Davide and Munro, Jonathan and Perrett, Toby and Price, Will and others},
  journal={IEEE Transactions on Pattern Analysis and Machine Intelligence},
  volume={43},
  number={11},
  pages={4125--4141},
  year={2020},
  publisher={IEEE}
}

@inproceedings{grauman2022ego4d,
  title={Ego4d: Around the world in 3,000 hours of egocentric video},
  author={Grauman, Kristen and Westbury, Andrew and Byrne, Eugene and Chavis, Zachary and Furnari, Antonino and Girdhar, Rohit and Hamburger, Jackson and Jiang, Hao and Liu, Miao and Liu, Xingyu and others},
  booktitle={Proceedings of the IEEE/CVF conference on computer vision and pattern recognition},
  pages={18995--19012},
  year={2022}
}

@article{zhang2023video,
  title={Video-llama: An instruction-tuned audio-visual language model for video understanding},
  author={Zhang, Hang and Li, Xin and Bing, Lidong},
  journal={arXiv preprint arXiv:2306.02858},
  year={2023}
}

@article{maaz2023video,
  title={Video-chatgpt: Towards detailed video understanding via large vision and language models},
  author={Maaz, Muhammad and Rasheed, Hanoona and Khan, Salman and Khan, Fahad Shahbaz},
  journal={arXiv preprint arXiv:2306.05424},
  year={2023}
}

@article{li2024llava,
  title={Llava-onevision: Easy visual task transfer},
  author={Li, Bo and Zhang, Yuanhan and Guo, Dong and Zhang, Renrui and Li, Feng and Zhang, Hao and Zhang, Kaichen and Zhang, Peiyuan and Li, Yanwei and Liu, Ziwei and others},
  journal={arXiv preprint arXiv:2408.03326},
  year={2024}
}

@inproceedings{yang2025egolife,
  title={Egolife: Towards egocentric life assistant},
  author={Yang, Jingkang and Liu, Shuai and Guo, Hongming and Dong, Yuhao and Zhang, Xiamengwei and Zhang, Sicheng and Wang, Pengyun and Zhou, Zitang and Xie, Binzhu and Wang, Ziyue and others},
  booktitle={Proceedings of the Computer Vision and Pattern Recognition Conference},
  pages={28885--28900},
  year={2025}
}

@inproceedings{rosetto2024castle,
    author = {Rossetto, Luca and Bailer, Werner and Dang-Nguyen, Duc-Tien and Healy, Graham and J\'{o}nsson, Bj\"{o}rn \TH{}\'{o}r and Kongmeesub, Onanong and Le, Hoang-Bao and Rudinac, Stevan and Sch\"{o}ffmann, Klaus and Spiess, Florian and Tran, Allie and Tran, Minh-Triet and Tran, Quang-Linh and Gurrin, Cathal},
    title = {The CASTLE 2024 Dataset: Advancing the Art of Multimodal Understanding},
    year = {2025},
    isbn = {9798400720352},
    publisher = {Association for Computing Machinery},
    address = {New York, NY, USA},
    url = {https://doi.org/10.1145/3746027.3758199},
    doi = {10.1145/3746027.3758199},
    booktitle = {Proceedings of the 33rd ACM International Conference on Multimedia},
    pages = {12629–12635},
    numpages = {7},
    location = {Dublin, Ireland},
    series = {MM '25}
}

@article{tulving1973encoding,
  title={Encoding specificity and retrieval processes in episodic memory.},
  author={Tulving, Endel and Thomson, Donald M},
  journal={Psychological review},
  volume={80},
  number={5},
  pages={352},
  year={1973},
  publisher={American Psychological Association}
}

@article{mcclelland1995there,
  title={Why there are complementary learning systems in the hippocampus and neocortex: insights from the successes and failures of connectionist models of learning and memory.},
  author={McClelland, James L and McNaughton, Bruce L and O'Reilly, Randall C},
  journal={Psychological review},
  volume={102},
  number={3},
  pages={419},
  year={1995},
  publisher={American Psychological Association}
}

@book{hunt2006distinctiveness,
  title={Distinctiveness and memory},
  author={Hunt, R Reed and Worthen, James B},
  year={2006},
  publisher={Oxford University Press}
}

@article{shaham2022stochastic,
  title={Stochastic consolidation of lifelong memory},
  author={Shaham, Nimrod and Chandra, Jay and Kreiman, Gabriel and Sompolinsky, Haim},
  journal={Scientific Reports},
  volume={12},
  number={1},
  pages={13107},
  year={2022},
  publisher={Nature Publishing Group UK London}
}

@inproceedings{subramaniam2026detecting,
  title={Detecting Social Engagement of Elderly From Lifelog Image-streams to Identify Effective Cues for Autobiographic Recall},
  author={Subramaniam, Vengateswaran and Subbaraju, Vigneshwaran and Roy, Debaditya and Krishna, Pramath and Kandappu, Thivya and Xu, Qianli},
  booktitle={Proceedings of the IEEE/CVF Winter Conference on Applications of Computer Vision},
  pages={3380--3389},
  year={2026}
}

@article{tian2026ego,
  title={Ego-R1: Agentic Chain-of-Tool-Thought for Ultra-Long Egocentric Video Reasoning},
  author={Tian, Shulin and Wang, Ruiqi and Guo, Hongming and Wu, Penghao and Dong, Yuhao and Wang, Xiuying and Yang, Jingkang and Zhang, Hao and Zhu, Hongyuan and Liu, Ziwei},
  journal={IEEE Transactions on Pattern Analysis and Machine Intelligence},
  year={2026},
  publisher={IEEE}
}
}
\appendix
\section{Query-Aware Captioning}
\label{supp:meta_captioning}
To improve the extraction of interaction records, we  define set of meta-questions based on EgoLifeQA~\cite{yang2025egolife}. Conditioning caption creation on these meta-questions enables the model to generate more comprehensive and retrieval-focused descriptions that retain details regarding participants, actions, objects, places, temporal context, and user intentions, hence enhancing memory quality for long-term question answering.
\subsection{Meta-Question Taxonomy}
\begin{figure}[!htbp]
  \centering
\includegraphics[width=\columnwidth]{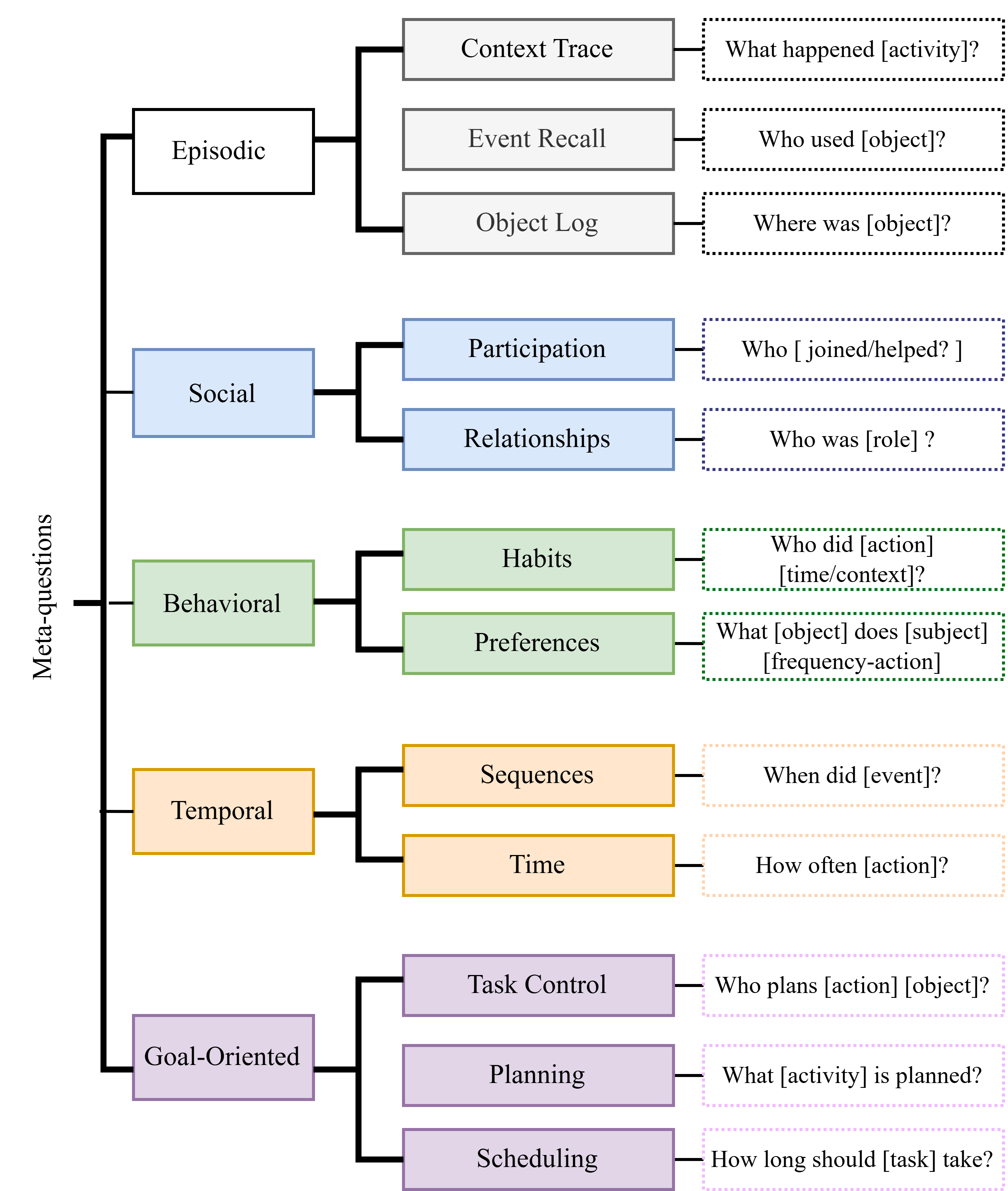}
  \caption{Meta-question taxonomy used for query-aware caption generation.}

  \label{fig:dendogram}
\end{figure}

To guide caption generation toward retrieval-relevant content, we organize 64 representative meta-questions into five functional categories — Episodic, Social, Behavioral, Temporal, and Goal-Oriented — each capturing a distinct facet of long-horizon egocentric reasoning targeted by EgoLifeQA~\cite{yang2025egolife}, as shown in Figure~\ref{fig:dendogram}. The meta-questions are not explicitly posed to the captioning model for response. Rather, they are integrated as implicit guidance during caption generation, prompting the model to assimilate information regarding participants, actions, objects, places, temporal context, and objectives that may subsequently be necessary for memory retrieval and question responding.

\begin{figure}[!htbp]
  \centering
  \includegraphics[width=\columnwidth]{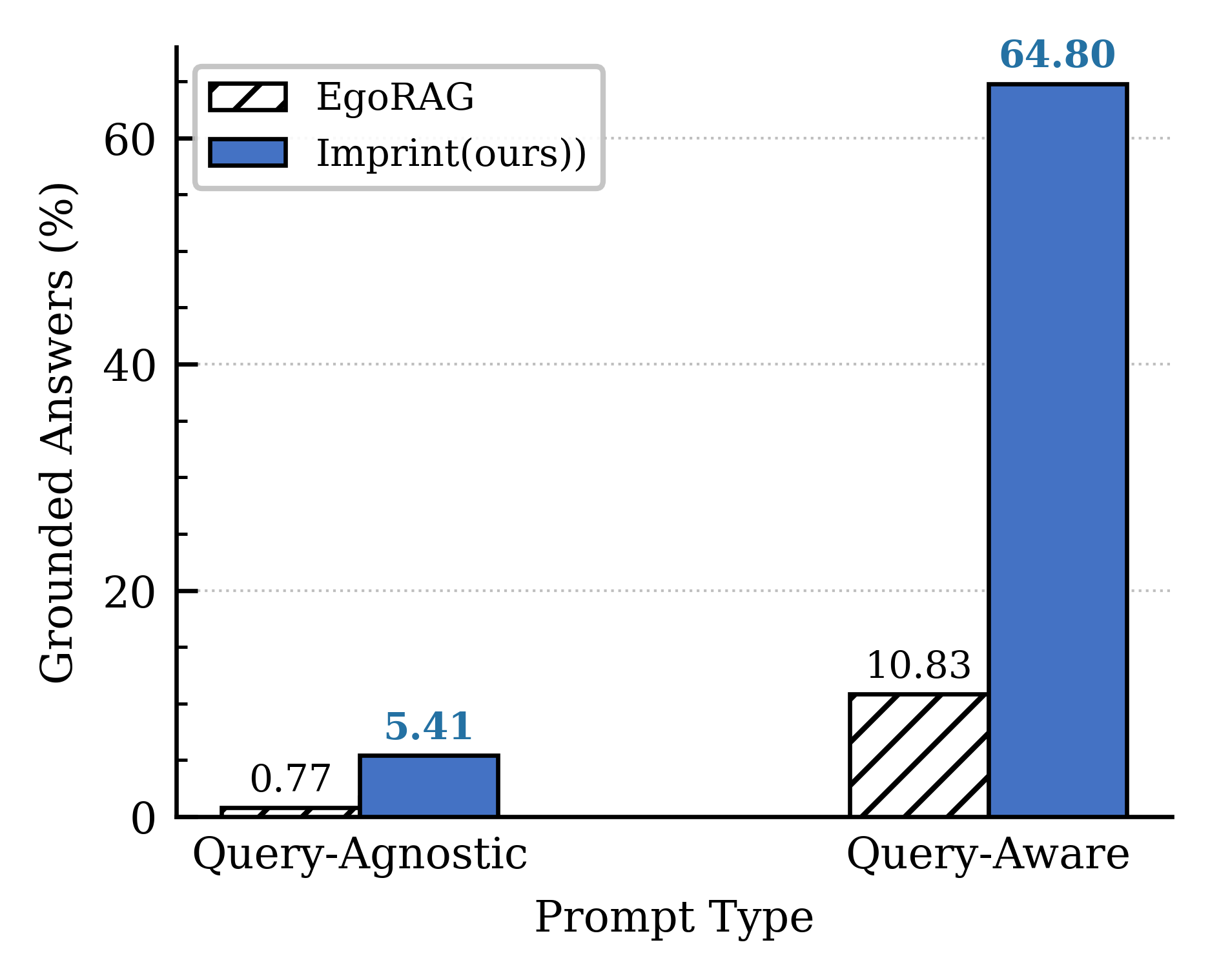}
  \caption{Query-aware caption generation improves retrieval grounding. Incorporating meta-question guidance during caption generation increases the proportion of grounded answers for both EgoRAG~\cite{yang2025egolife} and Imprint, with the largest gains observed for Imprint.}
  \label{fig:captioning-ablation}
\end{figure}
\subsection{Query-Aware Captioning Prompt}
\label{supp:captioning_prompt_des}
Figure~\ref{fig:prompt_caption} shows the full prompt used to generate a caption for each video segment. The prompt guides EgoGPT~\cite{yang2025egolife} to produce a first-person narrative using 'I' as the subject while leveraging the meta-questions as implicit guidance rather than explicit questions to answer.
\begin{figure*}[!htbp]
\small
\begin{tcolorbox}[
colback=white,
colframe=black,
title=Captioning Prompt,
boxrule=0.5pt,
arc=1mm
]
\begin{lstlisting}  
Imagine you are the character in the video, describing from a first-person perspective what you saw, and everything that happened over time. Use I as the subject.
Here is a set of 64 guiding questions to help you understand what details to include:
"
Who used the [object]? Where was the [object]? When was the [object/item/event] [action/state]? 
Who moved or handed the [object]? Who made the [object]? Who bought the [item]? 
What was the [object] used for? How many/how much was [quantity/currency/event]? Whose [object] is it?
What was in the [container]? Who did [action] with the [object]?
What did the [group] last use the [appliance] to bake? Who was the first to [action] the [object]? What was the last thing [person] used the [app/tool/device] for?
What was mentioned when [topic] was discussed, and by whom?
What was displayed on the [screen/device] when [event]?
Where is the [object], or what replaced it during [event/context]?
What was [person] washing with the [tool] last time?
What is the first [item] in the [container] yesterday?

Who joined or helped [person/group] during [activity/context]?
Who was/is [person] in the context of [role/context/timeframe]?
Who did [person] [action] during [event/time/context]?
Who acted or responded [first/already/later] during [event]?
Who was present or observed during [event or conversation]?
Who was involved in [location/time-based action]?

What did [subject] do or what happened during the last [event/context]?
What is/was the [object] used for or related to?
Who performed an action or was involved during [event/context]?
When did [event/action] happen or when did [subject] do [action]?
How did [subject] [action]?
How many times did [subject] [action]?
Where was [subject] or where did they interact with the [object]?

Who [habitual action] on [object]?
What does [subject] usually [action] after [event/action]?
Who [action] at/when [time/context]?
Who [action] the most [context]?
What [item/style/brand] does [subject] like/love/dislike [action]?

What [object] does [subject] [frequency] [action] [context]?
What is [subject] main responsibility in [context]?
What do [subject] [frequency] [action] on [object] [time]?
What is [subject] doing near [object] at [time]?
Where will [subject] never [action]?
Who is always the first [action] [object]? Which [object] does [subject] usually use [action]?
Where did [subject] [action] most frequently in [time]?
Who does [subject] usually [action] with? Why does [subject] [action] [object/context]?
How do [subject] [action] after [event]?
Who was near [subject] the last time [subject] [action]?
How do [subject] usually [action] [object]?

Who plans [action] [object]? What should/might [subject] [action] [time/context]?
What [items/tasks] are [time]?
What are/do [subject] planning/plan [action] for/in [time/meal]?
What [items/tasks] are needed [action] with/in [object/location]?
What do/are [subject(s)] planning [action]?
Why is/do [subject] [context]? How long should [task] take?
What [items] are still/was not [action] [context]?
How many [items] do/did [subject] [action/need]?
What does [subject] plan [action] [object/time]?
What are [subject] [action] for [meal/time]?
What does [subject] need [action] for [context]?
What [task] do/does [subject] have [time]?
What is [object] used for in the plan on [time]?
Where does [subject] plan [action] [context]?
What should [subject] do after [event]? When does [subject] plan [action]?
What is [subject] [action] [time]?
What is [subject]s [task]? What [item] can [subject] try at [event]? 
What should be [action] in [location]?
What [item] are [subject] [action] on [time]?
What tasks does [subject] still need [action]?

"Use these meta-questions only as hidden guidance to make your narration more detailed and complete.
Do NOT explicitly list the questions or provide separate answers. Instead, produce a smooth, natural, and continuous story-like caption that implicitly covers all the relevant details.
"""
\end{lstlisting}
\end{tcolorbox}
\caption{Query-aware captioning prompt with meta-questions used during caption generation.}
\label{fig:prompt_caption}
\end{figure*}
\subsection{Query-Aware Captioning Ablation}
Figure~\ref{fig:captioning-ablation} shows that query-aware captioning consistently enhances downstream QA performance for both caption-based and interaction-based retrieval methods. The enhancement is notably significant for Imprint, with Grounded Accuracy rising from 5.41\% to 64.80\%, indicating that preserving retrieval-relevant details during memory creation improves downstream retrieval and reasoning.

\section{Interaction Record Extraction}
\label{supp:IR_extract}
Following query-aware caption generation, each caption is parsed into one or more interaction records. Each interaction record contains the core fields defined in Section~3.1: person, action, object, start time, and end time. In addition, we extract auxiliary metadata, including tools, applications, locations, speech content, and contextual attributes.
These fields are retained as contextual metadata and are used during the answering stage to improve interaction disambiguation and provide better-grounded answers.

Figure~\ref{fig:fio_prompt} shows the extraction prompt. The prompt defines a fixed JSON schema containing \texttt{persons}, \texttt{action}, \texttt{object}, \texttt{tool}, \texttt{app}, location fields, \texttt{speech\_content}, and \texttt{attributes}, and instructs the model to generate one Interaction Record per distinct interaction while preserving speech verbatim. To resolve first-person references, the camera wearer’s identity is prepended to the caption before extraction, enabling consistent participant identification.in the persons field correctly.
\begin{figure*}[!htbp]
\small
\begin{tcolorbox}[
colback=white,
colframe=black,
title=Interaction Records Extraction Prompt,
boxrule=0.5pt,
arc=1mm
]
\begin{lstlisting}
You are a structured information extraction engine for egocentric video captions.

Your task: extract ALL distinct human interactions from the caption and return them
as a JSON array.


OUTPUT FORMAT: A JSON array only. No explanation. No markdown fences.


SCHEMA FOR EACH INTERACTION RECORD:
{
  "persons":         string,
  "action":          string,
  "object":          string,
  "tool":            string,
  "app":             string,
  "source_location": string,
  "target_location": string,
  "speech_content":  string,
  "attributes":      object
}

RULES:
1. Output ONLY a valid JSON array. No text before or after.
2. Create ONE FIO per distinct interaction.
3. "persons" MUST list ALL named persons in the scene.
4. "speech_content" MUST include exact words when a person speaks.
5. "object" and "tool" are different.
6. Use null for missing string fields.
7. Do NOT invent information not present in the caption.
8. The caption is from an egocentric camera.
9. "I" refers to the camera wearer.


EXAMPLES:

Caption: "I looked at my phone, opened the timer, and set it to 10 seconds."

Output:
[
  {
    "persons": "Jake",
    "action": "looked at",
    "object": "phone",
    "tool": null,
    "app": null,
    "source_location": null,
    "target_location": null,
    "speech_content": "",
    "attributes": {}
  }
]

Now extract all interactions record from the following caption:
{caption}
\end{lstlisting}
\end{tcolorbox}
\caption{System prompt used to extract Interaction Records from each egocentric caption.}
\label{fig:fio_prompt}
\end{figure*}
\section{Memory Compression}
\label{supp:fio_cmp}
Algorithm~\ref{alg:memory_compression} computes an importance score $w(f_i)$ for each Interaction Record using frequency, recency, and distinctiveness. Low-importance records are discarded, while redundant records are consolidated into existing memory entries with temporal history preserved. The resulting records form the compressed memory $\mathcal{M}$, followed by a final recency re-scoring step.

\begin{algorithm}[t]
\caption{Memory Compression}
\label{alg:memory_compression}
\small
\begin{algorithmic}[1]
\Require Video captions $\mathcal{X}$; importance threshold $\tau$; decay coefficient $\lambda$; prototype-similarity threshold $\delta_g$; prototype cap $N_{\max}$
\Ensure Compressed memory $\mathcal{M}$
\State Initialize memory $\mathcal{M} \leftarrow \emptyset$ and event-prototype set $\mathcal{G} \leftarrow \emptyset$
\For{each caption $x_k \in \mathcal{X}$}
    \State Extract interaction records $\mathcal{F}_k$ from $x_k$ using an LLM
    \For{each interaction record $f_i \in \mathcal{F}_k$}
        \State Encode $e_i \leftarrow \mathrm{Encoder}(a_i, o_i)$ and form contextual embedding $\tilde{e}_i$ 
        \State Assign $f_i$ to nearest prototype $g_{n_i}\!\in\!\mathcal{G}$ if $\mathrm{sim}(\tilde{e}_i, g_{n_i}) > \delta_g$, else create a new prototype (subject to cap $N_{\max}$)
        \State Update recurrence count $\nu_{n_i}$ and last-seen time $T^{\mathrm{last}}_{n_i}$
        \State Compute importance score: 
        \begin{equation*}
        w(f_i)=
        \overbrace{\log(1+\nu_{n_i})}^{\text{frequency}}\cdot
        \overbrace{e^{-\lambda\Delta T}}^{\text{recency}}\cdot
        \overbrace{\tfrac{1}{1+\log(1+N)}}^{\text{distinctiveness}}
        \end{equation*}
        \If{$w(f_i) < \tau$}
            \State Discard $f_i$
        \ElsIf{$\exists\, m\in\mathcal{M}$ whose action--object representation and contextual embedding are similar to $(a_i, o_i, \tilde{e}_i)$}
            \State Merge $f_i$ into $m$, preserving temporal order and aggregating importance 
        \Else
            \State Add new memory entry $m \leftarrow (\langle f_i\rangle,\, w(f_i))$ to $\mathcal{M}$
        \EndIf
    \EndFor
\EndFor
\State \Return $\mathcal{M}$
\end{algorithmic}
\end{algorithm}
\section{Retrieval and Question Answering}
\label{supp:retrieval}
The evidence retrieval and answer generation algorithm is shown in Algorithm~\ref{alg:retrieval}.
Questions are decomposed into structured components, candidate Interaction Records preceding $t_q$ are ranked using entity similarity, action similarity, and importance score , and answers are resolved either directly from retrieved records or through LLM reasoning over the top-5 retrieved records depending on the question type.
\begin{algorithm}[t]
\caption{Retrieval and Question Answering}
\label{alg:retrieval}
\small
\begin{algorithmic}[1]
\Require Question $q$ of type $t$ with timestamp $t_q$; memory store $\mathcal{M}$;
         retrieval weights $\gamma_1,\gamma_2,\gamma_3$; top-$k$; match threshold $\theta$
\Ensure  MCQ answer $y \in \{A,B,C,D\}$
\State $(o_q,\, a_q,\, \mathit{intent},\, r_q) \leftarrow \textsc{LLM}(q)$
       \Comment{decompose into retrieval cues}
\State $\mathcal{M}_q \leftarrow \{\, m \in \mathcal{M} : t^{s}_m \le t_q \,\}$
       
\ForAll{$m \in \mathcal{M}_q$}
    \State $s_m \leftarrow \gamma_1\cos(e_{o_q},e_{o_m})
            + \gamma_2\cos(e_{a_q},e_{a_m})
            + \gamma_3\, w(m)$ 
\EndFor
\State $\mathcal{R} \leftarrow \textsc{TopK}(\mathcal{M}_q,\, s;\; k)$
\If{$r_q \neq \varnothing$}
    \State $\mathcal{R} \leftarrow \textsc{ApplyTemporal}(\mathcal{R},\, r_q)$
\EndIf
\If{$\mathcal{R} = \varnothing$}
    \State \Return $\textsc{LLM}(q,\, \varnothing)$
\EndIf
\If{$t = \textsc{EntityLog}$}
    \If{$\mathit{intent} = \textit{when}$}
        \State $\delta \leftarrow t_q - \mathcal{R}[0].t^{s}$;\; map $\delta$ to a temporal label
    \ElsIf{$\mathit{intent} \in \{\textit{who, where, what}\}$}
        \State read $\mathcal{R}[0].\{\mathit{person, location, object}\}$ and map to an option
    \EndIf
    \State \Return matched option if score $\ge \theta$, else $\textsc{LLM}(q,\, \mathcal{R}[0])$
\ElsIf{$t = \textsc{RelationMap}$}
    \State aggregate co-occurring persons across $\mathcal{R}$
    \State \Return best option if score $\ge \theta$, else $\textsc{LLM}(q,\, \mathcal{R}_{:5})$
\ElsIf{$t = \textsc{HabitInsight}$}
    \State \Return most frequent field value over $\mathcal{R}$ if score $\ge \theta$,
           else $\textsc{LLM}(q,\, \mathcal{R}_{:5})$
\Else 
    \State \Return $\textsc{LLM}(q,\, \mathcal{R}_{:5})$
           
\EndIf
\end{algorithmic}
\end{algorithm}
\UseRawInputEncoding 
\begin{figure*}
\small
\begin{tcolorbox}[
colback=white,
colframe=black,
title=Query Decompose Prompt,
boxrule=0.5pt,
arc=1mm
]

\begin{lstlisting}
```
You are an information extraction engine for egocentric video question answering.

Your task is to convert the given natural-language question into a JSON object
that STRICTLY follows the schema below.

IMPORTANT RULES:
1. Output ONLY valid JSON. No explanations, no comments, no markdown.
2. Use ONLY the allowed values listed in the schema.
3. If a value is unknown or not explicitly stated, use null.
4. Do NOT invent new fields or values.
5. Choose the entity the question is ASKING ABOUT, not merely mentioning.
6. The "action" field: set ONLY if the question explicitly states a concrete action.
7. For contextual questions ("Now X does Y, what/who...?"), extract entity and action
8. "whose" and "who" both map to query_intent "who".

SCHEMA:
{
  "entity":           string | null,
  "entity_relation":  "reference" | "ownership" | "possession",
  "action":           string | null,
  "object_relation":  "contained_object" | "on_object" | null,
  "related_person":   string | null,
  "temporal_relation":"first_time" | "last_time" | "before" | "after" |
                       "habitual" | "yesterday" | "today" | null,
  "query_intent":     "who" | "where" | "when" | "what" | "why" | "how" | null,
  "speech_required":  true | false
}

FIELD GUIDANCE:

temporal_relation:
  "first_time"  — "first", "originally", "first to"
  "last_time"   — "last", "most recent", "latest"
  "before"      — "before", "previously", "used to"
  "after"       — "after", "since", "following"
  "habitual"    — "usually", "always", "often", "typically", "never", "rarely",
                   "every time", "tend to"
  null          — single specific event, no temporal qualifier

query_intent:
  "who"   — who / whose / which person
  "what"  — what / which object / app / tool / food / activity
  "where" — where / which place / location
  "when"  — when / how long ago / what time
  "why"   — why / what is the reason / because of what
  "how"   — how / in what way / by what method / how many / how much
  null    — yes/no questions

speech_required:
  Set true when the answer is likely spoken by a person (plan, preference,
  reason, decision, or opinion), and false otherwise.
```

Example decompositions:
Question: "Who used the screwdriver first?"
```json
{
  "entity": "screwdriver",
  "entity_relation": "reference",
  "action": "used",
  "object_relation": null,
  "related_person": null,
  "temporal_relation": "first_time",
  "query_intent": "who",
  "speech_required": false
}
```

\end{lstlisting}

\end{tcolorbox}

\caption{Prompt for query decomposition.}
\label{fig:query_decompose_prompt}
\end{figure*}
\subsection{Answering Prompt}
The retrieved evidence set $R$ is passed to \textsc{Qwen2.5-7B-Instruct} using the prompt shown in Figure~\ref{fig:ans_prompt}. The prompt provides the original question and its decomposed sub-questions, the inferred query intent, the four answer choices, and the ranked interaction memories retrieved from the compressed memory $\mathcal{M}$. The model is instructed to select a single answer option (A--D) and provide a brief evidence-grounded explanation.
\begin{figure}[t]
\small
\begin{tcolorbox}[
colback=white,
colframe=black,
title=Answering Prompt,
boxrule=0.5pt,
arc=1mm
]

\begin{lstlisting}
Given the decomposed question: {json.dumps(qwen_q_json)},
and the query intent: {qwen_q_json.get("query_intent")},
and the question info: {json.dumps(question_info)},
and question info contains choices:
A. {question.get("choice_a")}
B. {question.get("choice_b")}
C. {question.get("choice_c")}
D. {question.get("choice_d")}

Using the following first-person video evidence,
infer the most plausible answer.

Evidence (ranked):
{json.dumps(ranked, indent=2, default=str)}

Return response in one sentence and choose from A, B, C, D only.
Use the format:
Answer: <A/B/C/D>
Explain your choice in one sentence.
\end{lstlisting}

\end{tcolorbox}

\caption{LLM fallback prompt for multiple-choice answer generation over retrieved Interaction Records}
\label{fig:ans_prompt}
\end{figure}

\section{Additional Experiments}
\label{supp:additionale_exp}
In this section, we provide additional analyses covering stage-wise ablations over seven days, the effect of context history size, retrieval score ablations, ablation on $\mathrm{TopK}$ retrieval, EgoRAG's~\cite{yang2025egolife} hierarchical summaries,  and qualitative examples to further examine the behavior and robustness of the proposed memory framework.
\subsection{Per-Category Performance on EgoLifeQA}
\label{supp:detailed_category_result}
Table~\ref{tab:detailed_benchmark_results} shows the per-category Grounding Accuracy (GA) and QA Accuracy results summarized in Figure~3 of the main paper across the five EgoLifeQA~\cite{yang2025egolife} question types. Imprint improves grounding in all categories, with the largest gains observed for EntityLog,  EventRecall, RelationMap, and HabitInsight.

\begin{table*}[t]
\centering
\scriptsize
\setlength{\tabcolsep}{10pt}
\begin{tabular}{lcccccc}
\toprule
\textbf{Method} 
& \textsc{EntityLog} 
& \textsc{EventRecall} 
& \textsc{RelationMap} 
& \textsc{HabitInsight} 
& \textsc{TaskMaster} 
& \textbf{Overall(GA/Acc.)} \\
\midrule
\makecell[l]{EgoRAG~\cite{yang2025egolife}\\(Qwen 2.5 7B)}     
& 9.52 / 35.2 & 14.71 / 27.8 & 2.38 / \textbf{35.2}& 16.67 / 23.0 & 19.05 / 28.6 & 10.83 / 31.0 \\

\textbf{Imprint}
& 85.42 /\textbf{38.40}  & \textbf{61.22} / \textbf{38.89} & \textbf{63.64} / \textbf{35.20} & \textbf{80.00} / \textbf{24.59} & \textbf{21.74} / \textbf{36.51} & \textbf{64.80} / \textbf{35.80} \\
\bottomrule
\end{tabular}
\caption{Category-wise comparison of Grounded Accuracy (GA) and Accuracy (Acc.) across EgoRAG and Imprint on EgoLifeQA~\cite{yang2025egolife}.}
\label{tab:detailed_benchmark_results}
\end{table*}

\subsection{Stage-wise Ablation Over Seven Days}
We track the stage-wise ablation as the recording horizon grows from Day 1 to Day 7 to
verify that each compression stage contributes consistently and that the benefits do not
arise from compression alone. Table~\ref{tab:s_consolidation} shows GA / Acc. per configuration for each
cumulative day.
\begin{table*}[!htbp]
\centering
\scriptsize
\setlength{\tabcolsep}{12pt}
\begin{tabular}{lccccccc}
\toprule
\textbf{Configuration} & Day1 & Day1-2 & Day1-3 & Day1-4 & Day1-5 & Day1-6 & Day1-7 \\
\midrule
w/o Event-Prototype Assignment   & 62.5/\textbf{31.4} & 59.3/33.5 & 52.2/36.6 & 53.3/31.8 & 59.0/33.1 & 57.8/32.0 & 58.6/31.4 \\
w/o Interaction Importance       & 60.0/29.4 & 58.9/34.8 & 63.9/33.7 & 57.7/33.6 & 55.2/35.8 & 57.6/34.3 & 57.3/34.2 \\
w/o Interaction Consolidation    & 64.3/27.5 & 59.6/32.3 & 62.9/36.2 & 51.8/34.5 & 56.2/33.8 & 51.5/33.5 & 58.3/32.6 \\
Interaction Consolidation only   & \textbf{64.5}/30.4 & 58.2/34.2 & 56.0/34.1 & 54.4/34.5 & 52.5/34.8 & 54.7/33.5 & 56.2/32.4 \\
\textbf{Imprint}          & 56.2/\textbf{31.4} & 61.4/35.4 & 64.8/37.0 & 54.4/34.5 & 60.1/35.3 & 54.9/33.7 & \textbf{64.8/35.8} \\
\bottomrule
\end{tabular}
\caption{Stage-wise ablation on EgoLifeQA~\cite{yang2025egolife}  across cumulative day horizons. Each cell reports \textit{Overall GA/Acc} (\%); Imprint uses all stages.}
\label{tab:s_consolidation}
\end{table*}

\subsection{Ablation on Historical Context Parameters}
\label{supp:history_size_k}
Interaction records are contextualized using a short history of preceding interactions before prototype assignment and consolidation. The history window provides additional temporal context that helps distinguish semantically similar actions occurring in different situations.
Results for different history lengths are reported in Table ~\ref{tab:history-window}. Increasing the history size from 3 to 5 improves Grounded Answers, showing that more local interaction context helps retrieval. Increasing history size to 10, however, reduces Grounded Answers, suggesting that longer histories might introduce less relevant contextual information. We follow this trend and use k = 5 for all experiments.
The results show that the local interaction context is useful for memory formation, but most of the retrieval-relevant information is contained within a relatively short temporal neighborhood.
\begin{table}[!htbp]
\centering
\small
\setlength{\tabcolsep}{5pt}
\begin{tabular}{lc}
\toprule
Previous Interactions ($k$) & GA/Acc. \\
\midrule
3 & 51.75/28.60 \\
5 & \textbf{64.80}/\textbf{35.80} \\
10 & 53.06/29.40 \\
\bottomrule
\end{tabular}
\caption{Effect of the number of previous interactions $k$ used for contextualization during memory formation. Entries are GA / Acc.\ (\%); best at $k=5$.}
\label{tab:history-window}
\end{table}
\begin{table}[!htbp]
\centering
\small
\setlength{\tabcolsep}{10pt}
\begin{tabular}{lc}
\toprule
$\delta_h$ & GA/Acc. \\
\midrule
0.2 & 57.63/35.40\\
0.3 & \textbf{64.8}/\textbf{35.80} \\
0.5 & 58.14/34.40 \\
0.7 & 57.95/35.20 \\
\bottomrule
\end{tabular}
\caption{History-similarity threshold $\delta_h$. Only recent interactions whose similarity exceeds $\delta_h$ contribute context during event-prototype assignment. Entries are GA / Acc.\ (\%).}
\label{tab:history-simDelH}
\end{table}

\subsection{Ablation on Retrieval score }
\label{supp:gamma}
To analyze the contribution of each term in Eq.~3 (Section~3.3), we assigned a weight of zero to that term  and re-normalized the remaining weights and re-ran retrieval as shown in Table~\ref{tab:scoring-ablation}. Removing any component reduces grounding accuracy (GA) compared to the full score (64.8\%), with the largest drop observed for action similarity (48.2\%), followed by importance weighting (51.3\%) and entity similarity (52.7\%). This indicates that action cues are particularly important for distinguishing interactions involving similar objects, while entity similarity and importance score provide complementary retrieval signals.

\begin{table}[!htbp]
\centering
\scriptsize
\begin{tabular}{lccccc}
\toprule
\textbf{Variant} & $\gamma_1$& $\gamma_2$& $\gamma_3$ & \textbf{GA/Acc.} \\
\midrule
w/o EntitySim ($\gamma_1=0$) & 0 & 0.62 & 0.38 & 52.70/29.6 \\
w/o ActionSim ($\gamma_2=0$) & 0.75 & 0 & 0.25 & 48.25/28.6 \\
w/o Importance score ($\gamma_3=0$) & 0.65 & 0.35 & 0 & 51.30/30.8 \\
All& 0.53 & 0.29 & 0.18 & \textbf{64.80}/35.80\\
\bottomrule
\end{tabular}
\caption{Contribution of entity similarity, action similarity, and importance score to retrieval performance. Remaining weights are re-normalized after removing each component.}
\label{tab:scoring-ablation}
\end{table}
\subsection{Qualitative Analysis}
\label{supp:quali}
\begin{figure}[!htbp]
  \centering
  \includegraphics[width=\columnwidth]{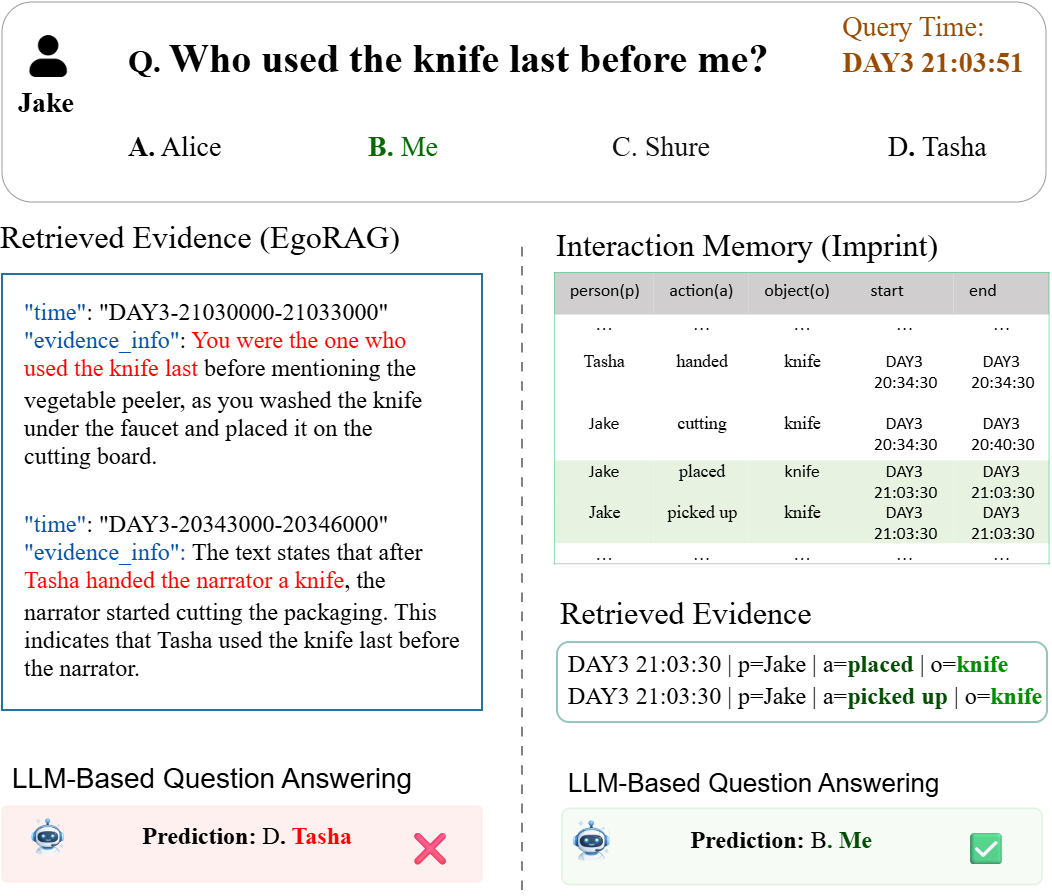}
  \caption{Retrieved two contradicting evidence, shown in red on the left, while Imprint timestamp-ordered and disambiguated cleanly: Tasha handed the knife at 20:34:30 $\rightarrow$. Jake began cutting at 20:34:30 $\rightarrow$ and placed the knife at 21:03:30 $\rightarrow$. Jake picked up knife at 21:03:30.}
  \label{fig:qa_entity}
\end{figure}

\paragraph{Temporal disambiguation (Figure~\ref{fig:qa_entity}).}
In case of contradictory evidence, caption-based retrieval retrieves both without preserving their temporal order. Imprint instead retrieves the relevant interaction records and orders them chronologically, resolving the apparent conflict cleanly:
\textit{Tasha handed the knife} ($20{:}34{:}30$) $\rightarrow$
\textit{Jake started cutting} ($20{:}34{:}30$) $\rightarrow$
\textit{Jake placed the knife} ($21{:}03{:}30$) $\rightarrow$
\textit{Jake picked up the knife} ($21{:}03{:}30$).

\begin{figure}[!htbp]
  \centering
  \includegraphics[width=\columnwidth]{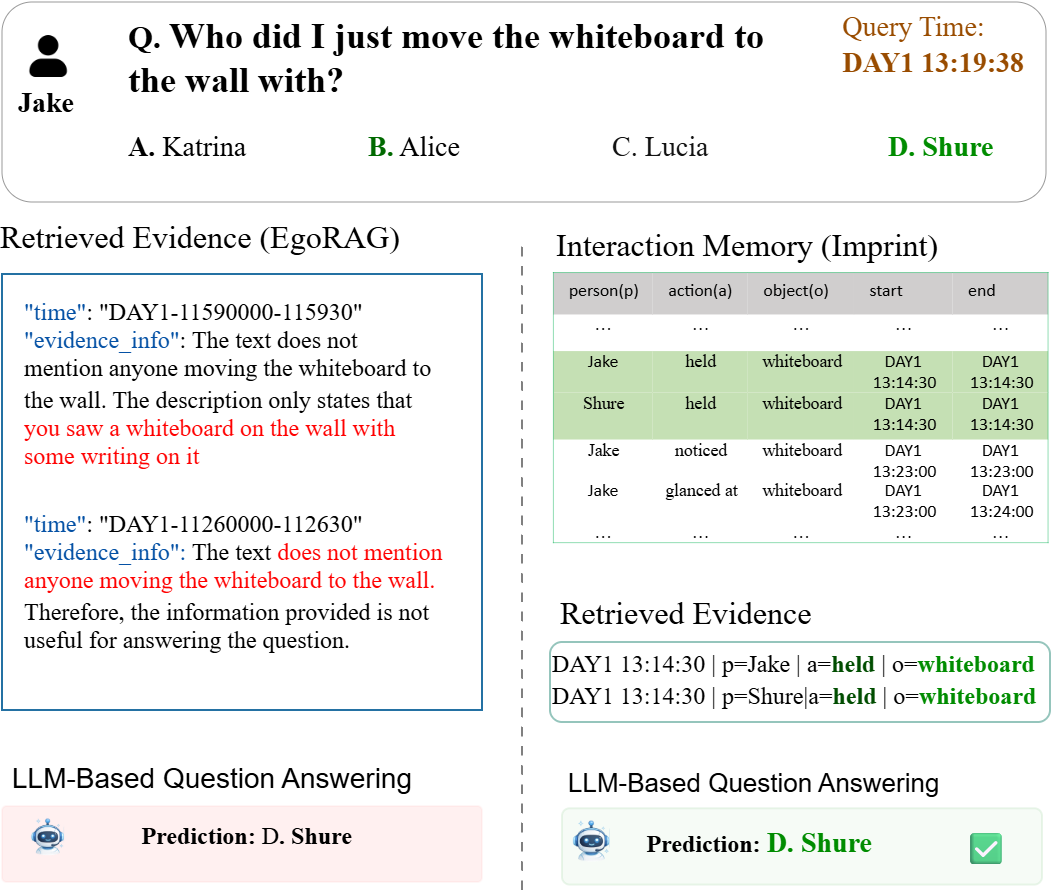}
 \caption{Retrieved evidence for a co-action query. Imprint retrieves the relevant interaction (\textit{Jake--Shure--whiteboard}) at the correct time, while EgoRAG~\cite{yang2025egolife} answers correctly but fails to retrieve supporting evidence for its prediction.}
  \label{fig:qa_relationMap}
\end{figure}

\paragraph{Grounded co-action retrieval (Figure~\ref{fig:qa_relationMap}).}
For questions involving co-actions at a specific time, Imprint retrieves the grounding interaction evidence (e.g., \textit{Jake} and \textit{Shure} interacting with the whiteboard at the target timestamp) and answers directly from retrieved evidence. While EgoRAG~\cite{yang2025egolife} retrieves wrong evidence and produces the correct answer only through language priors rather than grounded retrieval.

\begin{figure}[!htbp]
  \centering
  \includegraphics[width=\columnwidth]{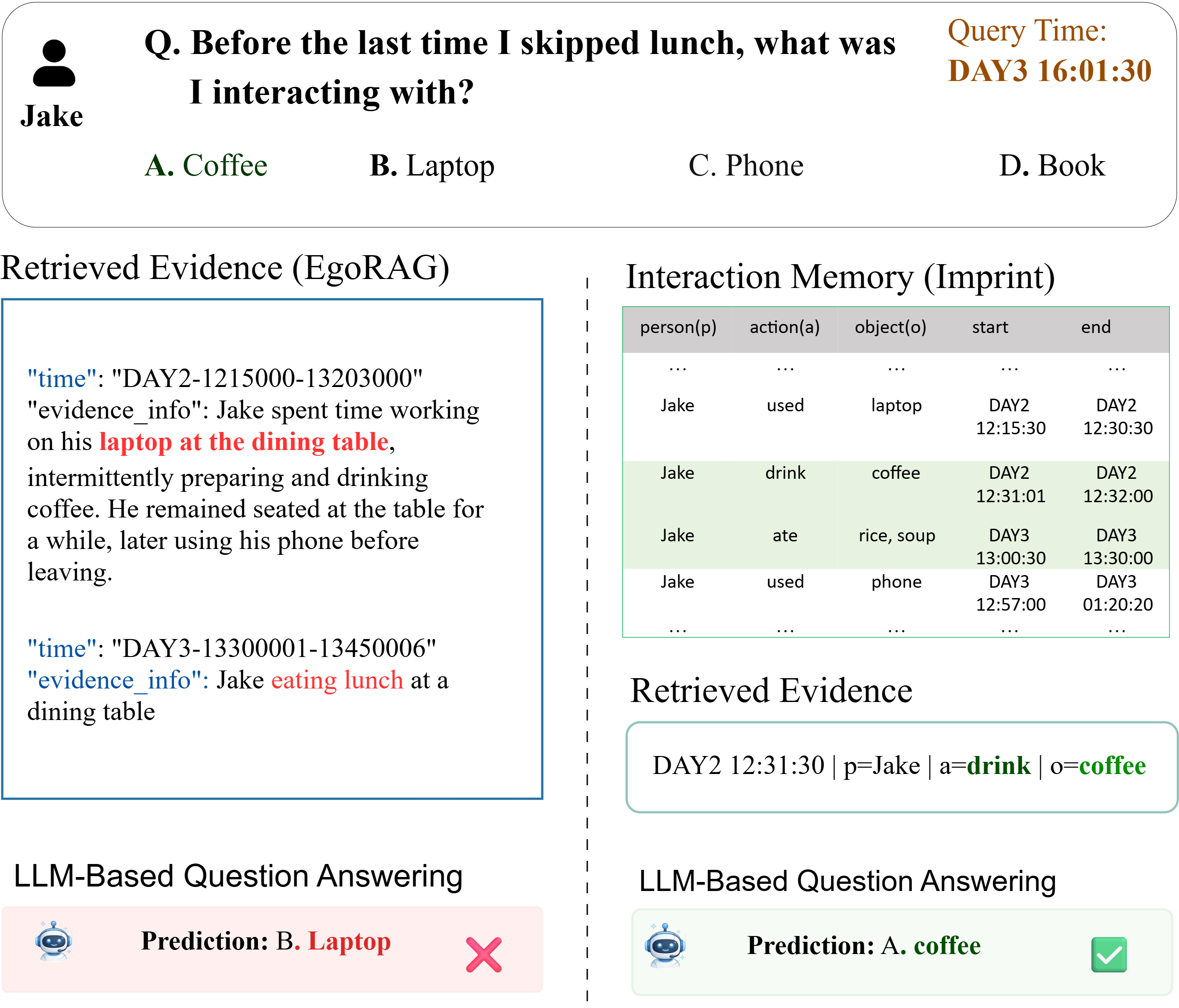}
 \caption{HabitInsight retrieval example. EgoRAG~\cite{yang2025egolife} retrieves an unordered evidence span and incorrectly relies on the dominant object (laptop) for prediction (left, red). Imprint preserves the temporal order of interactions, correctly identifying coffee as the activity immediately preceding the skipped lunch event, leading to the correct answer (A).}
  \label{fig:qa_lunch}
\end{figure}

\paragraph{Salient-object bias (Figure~\ref{fig:qa_lunch}).}
When retrieval returns broad caption spans containing multiple objects and activities, the language model often relies on the most prominent object mention rather than the interaction specified by the query. In this example, EgoRAG~\cite{yang2025egolife} retrieves evidence dominated by the activity ``working on a laptop'' and consequently predicts \emph{laptop}, overlooking the temporal constraint of the interaction immediately preceding the last skipped lunch. In contrast, Imprint preserves the chronological structure of interactions and retrieves the relevant sequence of events:
\textit{Jake used a laptop} ($\text{DAY2}~12{:}15{:}30$) $\rightarrow$
\textit{Jake drank coffee} ($\text{DAY2}~12{:}31{:}01$) $\rightarrow$
\textit{lunch skipped} $\rightarrow$
\textit{next eating event} ($\text{DAY3}~13{:}00{:}30$).
This temporal ordering correctly identifies \emph{coffee} as the interaction immediately preceding the skipped meal.
\subsection{EgoRAG's Hierarchical Summaries}
\begin{figure}[!htbp]
  \centering
  \includegraphics[width=\columnwidth]{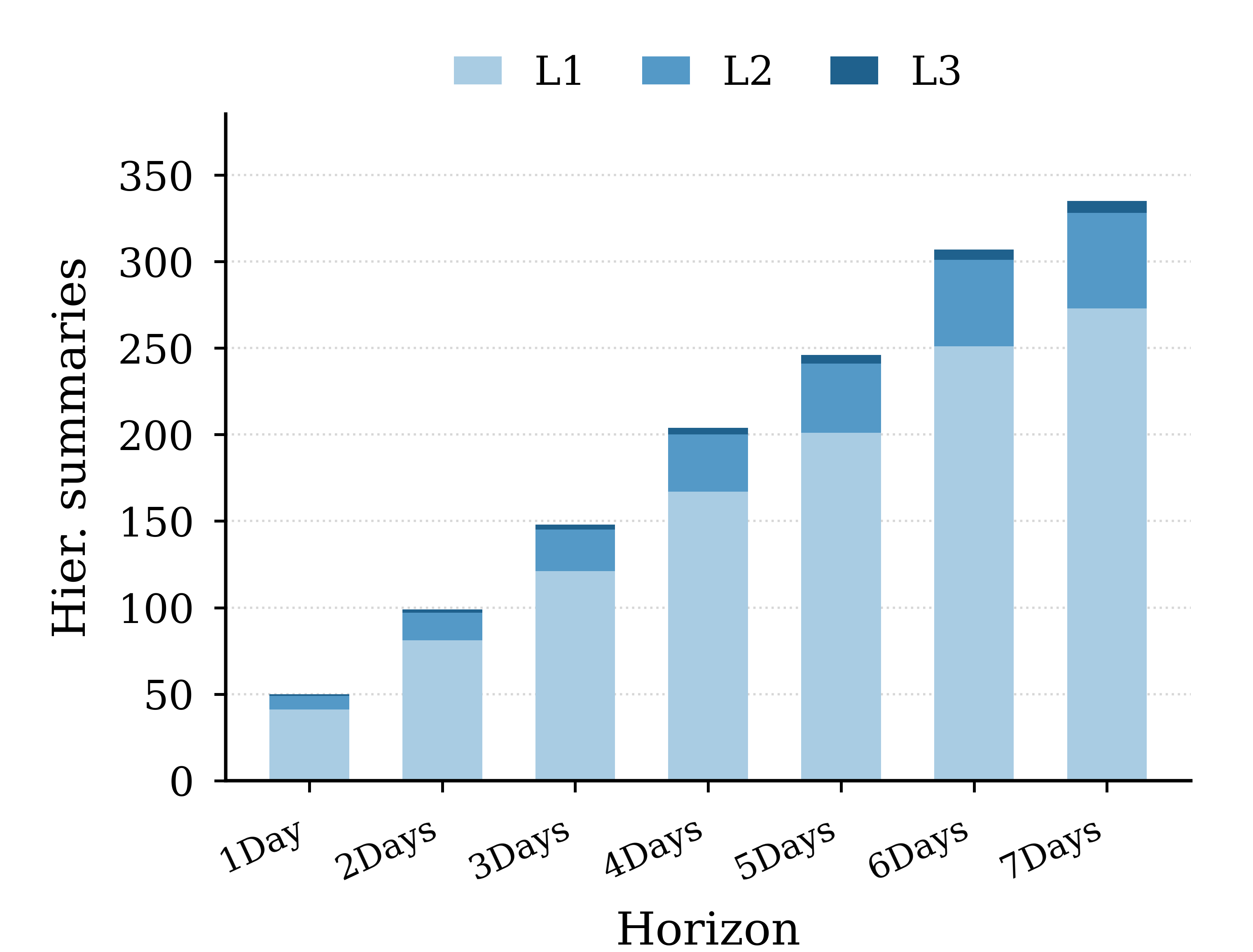}
  \caption{EgoRAG's~\cite{yang2025egolife} hierarchical summaries over seven days of recording, consisting of L1 (minute-level), L2 (hour-level), and L3 (day-level) summaries.}
  \label{fig:hier_summary}
\end{figure}

Figure~\ref{fig:hier_summary} shows EgoRAG's~\cite{yang2025egolife} hierarchical summary growing across the seven-day EgoLife recording. The number of summaries almost increases linearly  with recording duration, and most of these are L1 (minute-level) summaries. This makes storage and retrieval increasingly demanding as recording duration increases.
\label{supp:egorags's_summeries}

\subsection{Ablation on Top-K Retrieval }
We analyze the effect of the number of retrieved evidence $\mathrm{TopK}$ fed into the response LLM as shown in Table~\ref{tab:supp_topk}. GA is higher at $\mathrm{TopK}=5$ (64.8\%). Retrieving fewer evidence may miss essential supporting context, whereas retrieving more may introduce distracting information that weakens grounding and provides little benefit to QA accuracy.

\begin{table}[h]
\centering
\small
\setlength{\tabcolsep}{10pt}
\begin{tabular}{lc}
\toprule
$\mathrm{TopK}$ & GA/Acc. \\
\midrule
2 & 55.0/ 35.0\\
5 & \textbf{64.8}/\textbf{35.8} \\
10 & 61.8/35.6 \\
15 & 51.0/34.8 \\
20& 60.9/ 35.8\\
\bottomrule
\end{tabular}
\caption{Effect of the number of retrieved evidence (TopK) provided to the answering LLM. Entries report GA / Acc. (\%).}
\label{tab:supp_topk}
\end{table}
\end{document}